%% file: acl_latex.tex
\documentclass[11pt]{article}

\usepackage[final]{acl}

\usepackage{times}
\usepackage{latexsym}

\usepackage{graphicx}
\usepackage{subcaption}
\usepackage{enumitem}
\usepackage{amsmath}
\usepackage{amssymb}
\usepackage{bbm}
\usepackage{booktabs}
\usepackage{multirow}
\usepackage{multicol}
\usepackage{listings}
\usepackage{xcolor}

\usepackage{algorithm}
\usepackage{algpseudocode}

\usepackage{colortbl} 
\definecolor{gr}{gray}{0.95} 

\usepackage{tcolorbox}
\tcbuselibrary{skins,breakable}
\usepackage{listings}
\usepackage{xcolor}
\usepackage[T1]{fontenc}

\usepackage[utf8]{inputenc}

\usepackage{microtype}

\usepackage{inconsolata}

\usepackage{graphicx}

\title{Agentic Memory: Learning Unified Long-Term and Short-Term Memory Management for Large Language Model Agents}

\author{
  \textbf{Yi Yu\textsuperscript{1,2}},
  \textbf{Liuyi Yao\textsuperscript{2,\dag}},
  \textbf{Yuexiang Xie\textsuperscript{2}},
  \textbf{Qingquan Tan\textsuperscript{1}},
  \textbf{Jiaqi Feng\textsuperscript{1}},
\\
  \textbf{Yaliang Li\textsuperscript{2}},
  \textbf{and Libing Wu\textsuperscript{1,\dag}}
\\
  \textsuperscript{1}School of Cyber Science and Engineering, Wuhan University
\\
  \textsuperscript{2}Alibaba Group
\\
  \small{
   \texttt{\{yui1212,tanqingquan,jiaqiFeng,wu\}@whu.edu.cn}
  }
\\
  \small{
   \texttt{\{yly287738,yuexiang.xyx,yaliang.li\}@alibaba-inc.com}
  }
\\
 \small{
 \textbf{\dag Corresponding authors}
 }
}

\begin{document}
\maketitle
\input{sections/0_abstract}
\input{sections/1_intro}

\input{sections/2_relatedwork}

\input{sections/4_method}
\input{sections/5_experiment}
\input{sections/6_conclusion}

\section*{Acknowledgments}
This work was supported by the National Natural Science Foundation of China (62441237, U24A20336, 62272348, U22B2022), and Wuhan City Joint Innovation Laboratory for Next-Generation Wireless Communication Industry Featuring Satellite-Terrestrial Integration under Grant 4050902040448.

\section*{Limitations}

While AgeMem demonstrates strong performance across multiple settings, there remain opportunities for further extension. The current implementation adopts a fixed set of memory management tools, which provides a clear and effective abstraction but could be extended to support more fine-grained control in future work.

In addition, although we evaluate our approach on five representative long-horizon benchmarks and demonstrate zero-shot transfer across domains, these settings remain relatively controlled compared to open-ended real-world deployments. Evaluation in persistent, long-term dialogue or real-user interaction scenarios is an important next step, and we present our current study as establishing cross-domain transfer under controlled conditions as a foundation for such future work.

Finally, the training currently relies on HotpotQA as the source of three-stage trajectories; extending the curriculum to other data sources with richer interaction structures would further broaden the framework's applicability.



\bibliography{custom}
\newpage

\input{sections/appendix}

\end{document}

%% file: sections/0_abstract.tex
\begin{abstract}
Large language model (LLM) agents face fundamental limitations in long-horizon reasoning due to finite context windows, making effective memory management critical. Existing methods typically handle long-term memory (LTM) and short-term memory (STM) as separate components, relying on heuristics or auxiliary controllers, which limits adaptability and end-to-end optimization. In this paper, we propose Agentic Memory (AgeMem), a unified framework that integrates LTM and STM management directly into the agent’s policy. AgeMem exposes memory operations as tool-based actions, enabling the LLM agent to autonomously decide what and when to store, retrieve, update, summarize, or discard information. To train such unified behaviors, we propose a three-stage progressive reinforcement learning strategy and design a step-wise GRPO to address sparse and discontinuous rewards induced by memory operations. Experiments on five long-horizon benchmarks demonstrate that AgeMem consistently outperforms strong memory-augmented baselines across multiple LLM backbones, achieving improved task performance, higher-quality long-term memory, and more efficient context usage.
\end{abstract}

%% file: sections/1_intro.tex
\section{Introduction}

In long-horizon agentic tasks involving multi-step reasoning and complex workflows~\citep{chang2024agentboard}, the effectiveness of large language model (LLM) agents is fundamentally constrained by the information they can attend to at any given time, which we collectively refer to as the agent's \emph{memory}~\citep{xiong2025memory, goodyear2025effect}. 
Memory typically falls into two categories: long-term memory (LTM), which persistently stores user- or task-specific knowledge~\citep{zhong2024memorybank, jiang2024long}, and short-term memory (STM), which comprises the information contained in the current input context~\citep{wu2025human, gao2025efficient}. 
High-quality LTM supports efficient retrieval of accumulated knowledge, while effective STM management reduces redundancy and preserves salient context. Together, they mitigate the limitations of finite context windows, making their joint management crucial for improving agent performance in complex reasoning settings.
\begin{figure*}[!htp]
    \centering
    \includegraphics[width=\textwidth]{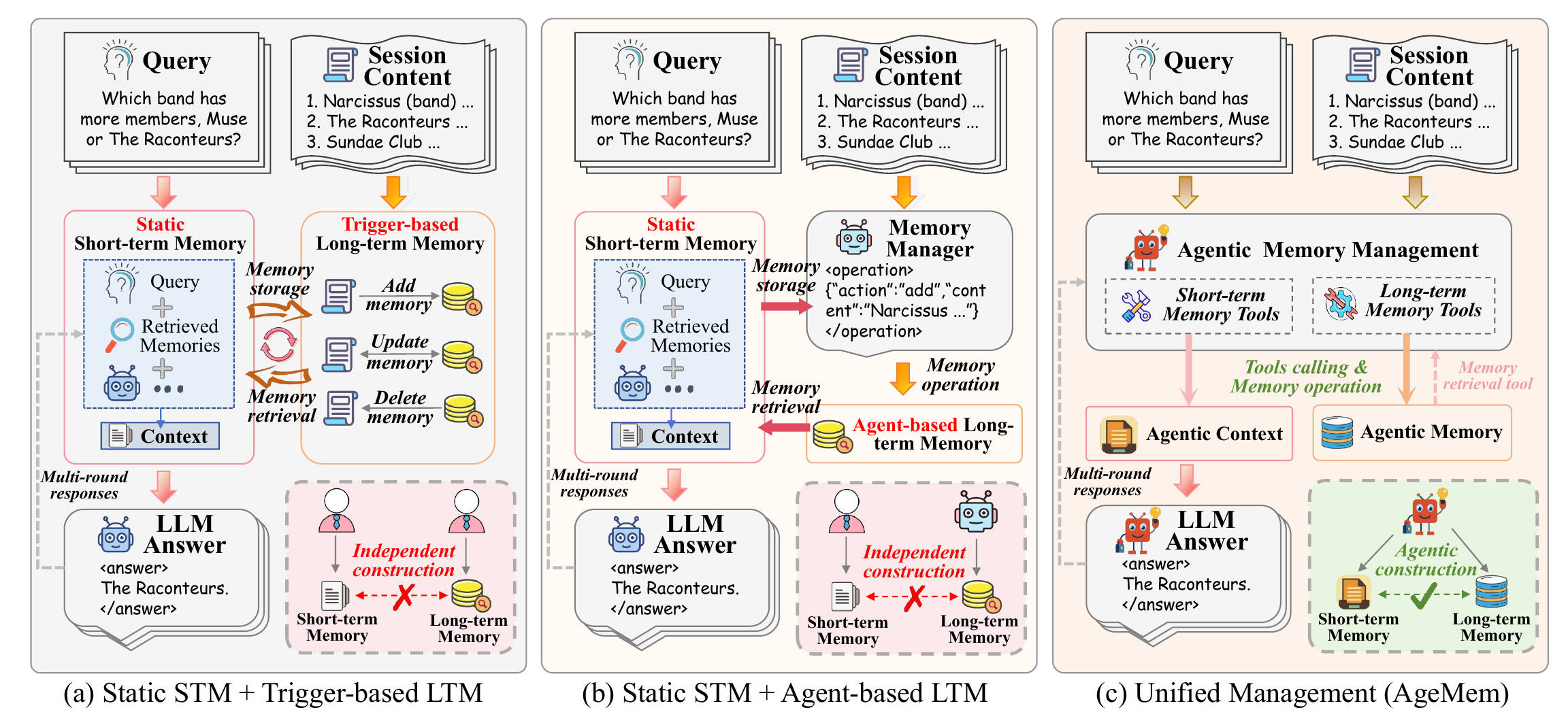}
    \vspace{-7mm}
    \caption{Comparison between independent and unified memory management frameworks. (Left) Traditional framework with static STM and trigger-based LTM. (Middle) Independent framework with an additional Memory Manager controlling LTM in an agent-based manner, while STM remains static. (Right) The proposed AgeMem framework, where LTM and STM are \textbf{jointly} and \textbf{intelligently} managed via explicit tool-based operations.}
    \vspace{-3mm}
    \label{fig:framework}
\end{figure*}

However, existing research has predominantly treated LTM and STM as independent components. 
STM is commonly enhanced through retrieval-augmented generation (RAG)~\citep{pan2025memory}, such as in MainRAG~\citep{chang2025main} and ReSum~\citep{wu2025resum}, which expand usable context via external retrieval or periodic summarization.
Although effective in some tasks, these methods rely heavily on predefined schedules or heuristic rules, 
potentially causing infrequent but critical details to be overlooked, while introducing unnecessary noise~\citep{ma2025should, dong2025survey}.
In contrast, LTM management has progressed along separate lines, typically categorized into \emph{trigger-based}~\citep{kang2025memory, wang2025mirix, wang2025inducing, chhikara2025mem0} and \emph{agent-based}~\citep{yan2025memory, hu2025evaluating, xu2025mem} paradigms.
The former executes fixed memory operations at predefined moments, whereas the latter incorporates a specialized memory manager to determine what and how to store. 
Despite offering more flexibility, most approaches still depend on handcrafted rules or auxiliary expert models, limiting adaptability and increasing system complexity~\citep{xiong2025memory}.

As a consequence, LTM and STM are typically treated as \emph{separate and loosely coupled modules}. 
As illustrated in Figure~\ref{fig:framework}, existing architectures generally follow two patterns: 
(a) static STM with trigger-based LTM, or 
(b) static STM with agent-based LTM. 
In both settings, the two memory systems are optimized independently and later combined in an ad hoc way, leading to fragmented memory construction and suboptimal performance in long-horizon reasoning tasks.
Thus, unifying the management of LTM and STM remains a necessary yet largely unexplored challenge.

Nevertheless, achieving unified memory management poses three fundamental challenges. (\textbf{C1}) \textbf{Functional heterogeneity coordination:} LTM and STM serve distinct yet complementary purposes: LTM determines what to store, update, or discard, while STM governs what to retrieve, summarize, or remove from the active context~\citep{zhang2025survey}. The challenge lies in designing a unified mechanism that orchestrates their interplay synergistically. (\textbf{C2}) \textbf{Training paradigm mismatch:} Existing reinforcement learning (RL) frameworks adopt markedly different training strategies for the two memory types~\citep{ma2024coevolving}. LTM-focused training often leverages session-level information available prior to interaction, whereas STM training typically injects distractors to simulate long-horizon contexts~\citep{sun2024llm}. Moreover, standard RL assumes continuous trajectories with stable rewards, which conflicts with the inherently fragmented and discontinuous experiences produced by memory operations~\citep{wu2025resum}, making end-to-end optimization particularly challenging. (\textbf{C3}) \textbf{Practical deployment constraints:} Many agent systems rely on an auxiliary expert LLM for memory control, significantly increasing inference cost and training complexity. How to integrate unified memory management directly into an agent without dependence on external expert models remains an open problem.

To address these challenges, we propose \textit{Agentic Memory (AgeMem)}, a unified framework that jointly manages LTM and STM, illustrated in Figure~\ref{fig:framework} (right). 
Unlike prior designs that treat memory as an external component, AgeMem integrates both memory types directly into the agent’s decision-making process. Through a unified tool-based interface, the LLM autonomously invokes and executes memory operations for both LTM and STM. 
Furthermore, we design a three-stage progressive RL strategy: the model first acquires LTM storage capabilities, then learns STM context management, and finally coordinates both forms of memory under full task settings. 
To address the fragmented experience issue across training stages, we design a step-wise Group Relative Policy Optimization (GRPO)~\citep{shao2024deepseekmath}, which propagates output rewards back to prior memory decisions, thereby alleviating the challenges posed by sparse and discontinuous rewards in RL. We evaluate AgeMem on five long-context, reasoning-intensive benchmarks. Comprehensive results show that AgeMem consistently outperforms strong baselines, validating the effectiveness of unified memory management.

Our main contributions are as follows:
\begin{itemize}[leftmargin=*, topsep=0pt, itemsep=0pt, parsep=0pt]
\item We propose \textit{Agentic Memory (AgeMem)}, a unified \textbf{agentic memory framework} that enables LLM-based agents to autonomously decide \textbf{when}, \textbf{what}, and \textbf{how} to manage both long-term and short-term memory.
\item We develop a \textbf{three-stage progressive RL strategy} equipped with a step-wise GRPO mechanism, facilitating effective end-to-end learning of unified memory management behaviors.
\item We conduct \textbf{comprehensive evaluations} across multiple models and long-horizon benchmarks, demonstrating the robustness and effectiveness of AgeMem in complex agentic tasks.
\end{itemize}

%% file: sections/2_relatedwork.tex
\section{Background and Related Work}
\textbf{Long-term memory (LTM).} 
Persistent LTM is crucial for LLM-based agents operating over extended horizons~\citep{wang2025karma, li2025hello}.
Recent work has explored diverse architectural designs for modeling LTM. 
LangMem~\citep{langmem2025} provides a modular framework that supports multiple memory types, while A-Mem~\citep{xu2025mem} adopts a Zettelkasten-inspired design that links structured knowledge units to facilitate consolidation. 
Mem0~\citep{chhikara2025mem0} proposes a scalable extract-update pipeline and extends it to a graph-based variant for structured reasoning. Zep~\citep{rasmussen2025zep} represents memory as a temporal knowledge graph to enable cross-session and time-aware reasoning. Although effective in organizing and retrieving information, these approaches largely rely on predefined memory structures or heuristic update rules. 
As memory grows, such designs commonly suffer from increased system complexity and lack adaptive, learning-based strategies for prioritization and forgetting.
In contrast, our work aims to learn an adaptive memory policy that allows agents to dynamically decide what to store, update, or forget, depending on task demands and long-term utility.
\\
\textbf{Short-term memory (STM).} 
STM in agentic LLMs primarily concerns context selection and retrieval~\citep{wang2024agent, jin2024llm}.
Retrieval-Augmented Generation (RAG)~\citep{pan2025memory, salama2025meminsight, kagaya2024rap} is the dominant paradigm, expanding usable context by injecting retrieved content into prompts.
While effective, RAG does not fundamentally prevent context explosion in long-horizon settings and may introduce irrelevant or distracting information.
To address this issue, ReSum~\citep{wu2025resum} periodically compresses interaction histories into compact reasoning states, allowing agents to operate beyond fixed context-window constraints. 
Yet its summarization schedule remains largely predefined, and aggressive compression risks discarding rare but crucial details.
Our approach instead enables agents to \emph{learn} when and how to retrieve, summarize, or filter context, achieving a more flexible balance between efficiency and information preservation.
\\
\textbf{Reinforcement learning for LLMs.} 
Reinforcement learning has become an effective paradigm for improving the decision-making and reasoning capabilities of LLM-based agents~\citep{yao2022react, jin2025search, qian2025toolrl, chaudhari2025rlhf}. 
Among recent advances, GRPO~\citep{shao2024deepseekmath} enhances stability by optimizing policies based on the relative quality of sampled trajectories, removing the need for an explicit value function. 
GRPO and its variants~\citep{gilabert2025terminology, wang2025grpo} have shown strong performance in complex reasoning tasks.
However, existing RL-based systems generally treat memory as a static or external component, making them ill-suited for the discontinuous and fragmented trajectories associated with memory operations~\citep{yan2025memory, zhang2025memory}. 
In contrast, our work integrates RL directly into the memory management process, enabling unified training of both language generation and memory operations.
\\
\textbf{Positioning relative to RL-based memory agents.}
Recent work~\citep{yan2025memory,zhang2025memory} models 
memory operations as actions and applies RL to optimize them, which 
shares surface similarity with our approach. However, these methods typically optimize \emph{one aspect of memory at a time} while treating retrieval, summarization, or short-term context handling as fixed heuristics or separately tuned modules. As a result, early storage decisions and later reasoning behavior are only loosely coupled, and the learning signal does 
not explicitly connect them. AgeMem instead formulates memory usage as a single learnable control problem under delayed supervision: a 
unified policy over heterogeneous memory actions (both persistent 
LTM operations and contextual STM operations) is trained end-to-end, so 
that storage, retrieval, filtering, and summarization decisions are all 
optimized jointly with respect to the same terminal task reward.

%% file: sections/4_method.tex
\section{Method}

We propose \textit{Agentic Memory (AgeMem)}, a unified memory framework that enables LLM agents to autonomously manage both LTM and STM in an end-to-end manner. 
As illustrated in Figure~\ref{fig:framework} (right), AgeMem integrates memory management capabilities directly into the agent via a set of specialized tools, enabling the model to learn optimal strategies for unified memory management through a three-stage progressive strategy.

\subsection{Problem Formulation}
\label{sec:problem_formu}


\textbf{Unified RL formulation for AgeMem.} At each time step $t$, the agent observes a state $s_t \in \mathcal{S}$ composed of the conversation context (short-term memory) $C_t$, the long-term memory store $\mathcal{M}_t$, and the task specification $\mathcal{T}$: $s_t = (C_t, \mathcal{M}_t, \mathcal{T})$.
The specification $\mathcal{T}$ includes the input query $q$, contextual information $I_q$, and (for training only) the expected answer $A_q$. This formulation enables the agent to ground its decision-making in both transient context and persistent knowledge.

Given $s_t$, the agent selects an action $a_t \in \mathcal{A}$ from a hybrid action space that includes language generation as well as memory operations. The decision is governed by a parameterized policy $\pi_\theta$, defined as 
$\pi_\theta(a_t | s_t) = P(a_t | s_t; \theta)$,
where $\theta$ denotes the LLM parameters and $a_t \sim \pi_\theta(\cdot|s_t)$.
For a trajectory $\tau = (s_1, a_1, \ldots, s_T, a_T)$, the cumulative reward is defined as:
\begin{equation}
R(\tau) = \sum_{i} w_i \cdot R_i\left(\tau\right) + P_{\text{penalty}}(\tau),
\end{equation}
where $R_i$ captures task performance and memory quality, and $P_{\text{penalty}}$ discourages redundant storage, excessive tool usage, and uncontrolled context expansion. The optimization objective is:
\begin{equation}
\theta^{*} = \arg\max_{\theta} \mathbb{E}_{\tau \sim \pi_\theta} [ R(\tau) ].
\end{equation}
This formulation treats memory management as an integral component of the agent’s policy, replacing handcrafted heuristics with a learnable mechanism.
\\
\textbf{Three-stage trajectory structure.} To capture long-horizon interactions and progressively train memory capabilities, each trajectory is divided into three consecutive stages: $\tau = (\tau^{(1)}, \tau^{(2)}, \tau^{(3)}),$
with a total length of $T = T_1 + T_2 + T_3$. In Stage~1, the agent engages in casual interactions and may store useful information into LTM. Stage~2 introduces distracting or irrelevant content, requiring the agent to manage its STM through selective retention and compression. Stage~3 presents a task that depends on coordinated use of both retained context and earlier accumulated LTM. A key aspect of this design is that the long-term memory $\mathcal{M}_t$ persists across all stages, allowing early knowledge to influence later decisions. In contrast, the context $C_t$ is reset between Stages~1 and 2 to prevent information leakage across phases. The reset before Stage~2 ensures the agent cannot solve the final task via residual context, thereby forcing proper retrieval from LTM and enabling effective training of memory operations.

At each step, we collect an experience tuple $e_t = (s_t, a_t, r_t, \log \pi_{\theta_{\text{old}}}(a_t | s_t))$,
where $r_t$ is typically zero for intermediate steps and assigned after trajectory completion, and $\log \pi_{\theta_{\text{old}}}(a_t | s_t)$ denotes the $log$ probability under the old policy $\pi_{\theta_{\text{old}}}$. This representation enables step-wise credit assignment under GRPO~\citep{shao2024deepseekmath} and allows the agent to attribute long-term rewards to specific memory decisions across stages. By structuring trajectories in this staged yet continuous manner, the agent learns temporally coherent and task-adaptive memory policies essential for robust long-horizon reasoning.

\begin{table}[t]
\centering
\caption{Memory management tools in AgeMem for manipulating long-term memory (LTM) and short-term memory (STM).}
\label{tab:tools}
\small
\begin{tabular}{@{}lll@{}}
\toprule
\textbf{Tool} & \textbf{Target} & \textbf{Function} \\
\midrule
\textsc{Add}      & LTM & Add new knowledge to $\mathcal{M}_t$ \\
\textsc{Update}   & LTM & Modify entries in $\mathcal{M}_t$ \\
\textsc{Delete}   & LTM & Remove entries from $\mathcal{M}_t$ \\
\midrule
\textsc{Retrieve} & STM & Retrieve entries from $\mathcal{M}_t$ to $C_t$ \\
\textsc{Summary}  & STM & Summarize segments in $C_t$ \\
\textsc{Filter}    & STM & Filter out irrelevant segments from $C_t$ \\
\bottomrule
\end{tabular}
\end{table}

\subsection{Memory Management via Tool Interface}

AgeMem exposes memory-related operations to the LLM agent through an explicit tool interface (Table~\ref{tab:tools}). The agent can modify its persistent LTM using \textsc{Add}, \textsc{Update}, and \textsc{Delete}, while exercising fine-grained control over STM through \textsc{Retrieve}, \textsc{Summary}, and \textsc{Filter}. Incorporating these tools into the action space transforms memory control from an external heuristic pipeline into an intrinsic component of decision-making. This design allows the agent to adaptively manage memory according to task structure, history, and context.

Each tool serves a distinct functional role in memory management. 
\textbf{LTM operations}: \textsc{Add} inserts a new entry into the long-term store $\mathcal{M}_t$; \textsc{Update} modifies an existing entry identified by \texttt{memory\_id}; \textsc{Delete} removes an entry from $\mathcal{M}_t$ to prevent accumulation of stale knowledge. 
\textbf{STM operations}: \textsc{Retrieve} brings the top-$k$ semantically relevant memories from $\mathcal{M}_t$ into the active context $C_t$; \textsc{Summary} compresses a specified span of interaction history into a concise representation, reducing context size while preserving essential information; \textsc{Filter} removes context messages whose semantic similarity to a given criterion exceeds a threshold $\theta_{f}$, suppressing irrelevant or 
distracting content. Together, these six operations provide the agent with expressive yet interpretable control over its memory lifecycle.
Full formal definitions and system prompts are provided in Appendix~\ref{app:tools}.

\subsection{Three-Stage Progressive RL Strategy}

To learn unified and stable memory behaviors, we propose a progressive three-stage training strategy. 
For each task instance $q \in \mathcal{T}$, the agent generates a complete trajectory:
\begin{equation}
\tau_k^{(q)} = \big(\tau_k^{(1)}, \, \tau_k^{(2)}, \, \tau_k^{(3)}\big), \quad k = 1, \dots, K,
\end{equation}
where $K$ denotes the number of independent rollouts, and each sub-trajectory $\tau_k^{(i)}$ corresponds to a specific training stage. 
\\
\textbf{Stage~1 (LTM construction).}  
The agent is exposed to contextual information $I_q$ in a casual conversational setting. 
The goal is to identify salient information and store it into LTM $\mathcal{M}_t$. 
During the interaction, the short-term context $C_t$ evolves naturally, and the agent may invoke LTM-related tools when appropriate.
Formally, this stage yields a sub-trajectory $\tau_k^{(1)} = \{e_t\}_{t=1}^{T_1}$,
where each experience tuple $e_t$ follows the definition in Section~\ref{sec:problem_formu}.
\\
\textbf{Stage~2 (STM control under distractors).}  
The short-term context is reset, while the constructed LTM $\mathcal{M}_t$ is retained. The agent is then presented with \emph{distractor messages}---natural-language 
utterances (e.g., questions or short statements) that resemble plausible conversational context but are deliberately unrelated to the target query---so that the agent must learn to filter or ignore them rather than rely on them for the final answer. 
The objective is to learn proactive STM control through tool-based operations, such as filtering or summarizing context, in order to suppress noise and preserve useful information.
This process forms the sub-trajectory $\tau_k^{(2)} = \{e_t\}_{t=T_1+1}^{T_1+T_2}$,
which emphasizes context filtering and compression capability.
\\
\textbf{Stage~3 (Integrated reasoning and memory coordination).}  
Finally, the agent receives a formal query $q$ requiring both accurate reasoning and effective memory retrieval. 
The agent must retrieve relevant knowledge from $\mathcal{M}_t$, appropriately manage the context $C_t$, and generate a final answer.
This stage produces $\tau_k^{(3)} = \{e_t\}_{t=T_1+T_2+1}^{T}$,
which evaluates the ability of the agent to coordinate long-term memory, short-term context management, and task solution in an end-to-end manner.

All three segments form a complete trajectory:
\begin{equation}
    \tau_k^{(q)}=(e_1, e_2, \ldots, e_T),\quad T = T_1 + T_2 + T_3,
\end{equation}
which is then used for policy optimization in the subsequent step-wise GRPO procedure. For a batch of $B$ tasks, we further aggregate all experiences from $K$ independent rollouts into a unified set $\mathcal{E} = \bigcup_{q=1}^{B} \bigcup_{k=1}^{K} \{ e_t \mid e_t \in \tau_k^{(q)} \}$, with a total size of $|\mathcal{E}| = B \times K \times \bar{T}$, where $\bar{T}$ denotes the average trajectory length. More detailed rollout processes are provided in the Appendix~\ref{app:algorithm}.
\\
\textbf{Generalizability of the three-stage curriculum.}
The three-stage structure is not tied to QA-style supervision; it requires only a temporal separation between information exposure and task execution so that the usefulness of memory decisions can be evaluated under delayed outcomes. Each stage fulfills a functional role that can be instantiated in diverse tool-using settings: \textbf{Stage~1} (information acquisition) requires any pre-task context from which the agent may selectively store information---this could be environment descriptions, retrieved documents, or prior dialogue history, not exclusively QA supporting facts. 
\textbf{Stage~2} (interference and context pressure) is already generated synthetically via \textsc{DistractorGen} (detailed in Appendix~\ref{app:algorithm}) and requires no dataset annotations. 
\textbf{Stage~3} (task execution) provides the downstream objective whose delayed reward determines whether earlier storage and filtering decisions were useful. In this sense, the curriculum depends on information timing and delayed credit assignment rather than QA-style factual supervision. 

\subsection{Step-wise GRPO for Unified Management}

We adopt a step-wise variant of GRPO to connect long-range task rewards with memory decisions across all stages.
For task $q$, let $G_q = \{\tau_1^{(q)},\ldots,\tau_K^{(q)}\}$ denote the group of parallel rollouts.
Each trajectory yields a terminal reward $r_T^{(k,q)} = R(\tau_k^{(q)})$. We compute the group-normalized advantage for the terminal step as:
\begin{equation}
    A_T^{(k,q)} = \frac{r_T^{(k,q)} - \mu_{G_q}}{\sigma_{G_q} + \epsilon},
\end{equation}
where $\mu_{G_q}$ and $\sigma_{G_q}$ are the mean and standard deviation of rewards within $G_q$, $\epsilon$ prevents division by zero. This advantage is then \emph{broadcast} to all preceding steps of the same trajectory $A_t^{(k,q)} = A_T^{(k,q)}$,
which assigns a consistent learning signal to all memory and reasoning actions along the trajectory, including those in Stage~1 and Stage~2. 
In doing so, the final task outcome supervises every intermediate memory decision, enabling long-range credit assignment across heterogeneous stages. We then augment the experience set with advantages, $\mathcal{E} = \bigcup_{q,k}^{B,K} \{ (e_t, A_t) | e_t \in \tau_k^{(q)},A_t=A_t^{(k,q)} \}$. 

Following GRPO, we maximize the expected objective over all experiences:
\begin{equation}
\begin{aligned}
    J(\theta) &= \mathbb{E}_{(e_t, A_t) \sim \mathcal{E}} \big[\rho_t A_t - \beta D_{\text{KL}}[\pi_{\theta}\|\pi_{\text{ref}}] \big] \\
    &= \frac{1}{|\mathcal{E}|} \sum_{q=1}^{B} \sum_{k=1}^{K} \sum_{t=1}^{T_k^{(q)}} \big[\rho_t^{(k,q)} A_t^{(k,q)} - \beta D_{KL}^{(k,q)} \big],
\end{aligned}
\end{equation}
where the importance ratio 
$\rho_t^{(k,q)} = \frac{\pi_\theta(a_t| s_t)}{\pi_{\theta_{\text{old}}}(a_t |s_t)}$ 
controls the update magnitude under the new policy, 
$D_{\text{KL}}^{(k,q)}$ denotes the KL divergence penalty between the current policy $\pi_\theta$ and a fixed reference $\pi_{\text{ref}}$, 
and $\beta$ is a coefficient that balances exploration and training stability.

\subsection{Reward Function Design}

We design a composite reward that evaluates both downstream task performance and the quality of memory management. The total trajectory-level reward is defined as
\begin{equation}
R(\tau) = \mathbf{w}^\top \mathbf{R} + P_{\text{penalty}},
\end{equation}
where $\mathbf{w} = [w_{\text{task}}, w_{\text{context}}, w_{\text{memory}}]^\top$ are tunable coefficients, and $\mathbf{R} = [R_{\text{task}}, R_{\text{context}}, R_{\text{memory}}]^\top$ correspond to rewards for task completion, context management, and long-term memory management. The penalty term $P_{\text{penalty}}$ captures violations such as context overflow or exceeding the interaction limit. Below, we summarize each component, and precise formulas are provided in the Appendix~\ref{app:rewards}.
\\
\textbf{Task completion reward $R_{\text{task}}$.}  
This term provides the primary learning signal by assessing whether the agent solves the task correctly. We obtain a scalar score using an LLM-based judge $S_{\text{judge}}(A_{\text{pred}}, A_q) \in [0,1]$, optionally applying a penalty when no answer is produced. This reward encourages accurate, complete task solutions and remains the dominant component to ensure alignment with task objectives.
\\
\textbf{Context management reward $R_{\text{context}}$.}  
This component evaluates STM behavior, focusing on how effectively the agent controls the active context $C_t$. It combines three factors: (i) compression efficiency, promoting economical token usage; (ii) preventive actions, rewarding early summarization or filtering to avoid overflow; and (iii) information preservation, penalizing the loss of critical query-related content. Each factor is normalized, allowing the reward to balance context efficiency against retention of essential information.
\\
\textbf{Memory management reward $R_{\text{memory}}$.}
This term evaluates LTM operations. It aggregates signals for: (i) storage quality, measured as the fraction of stored entries labeled as high-quality and reusable; (ii) maintenance, rewarding meaningful update or delete operations to mitigate memory staleness; and (iii) semantic relevance, computed using an LLM-based score between retrieved memories and the query. Together, these signals incentivize selective, high-value memory construction and responsible upkeep over time.
\\
\textbf{Penalty terms $P_{\text{penalty}}$.}  
Penalties discourage undesirable behaviors such as exceeding the maximum number of dialogue turns or triggering context overflow. Penalty coefficients are chosen so that such violations lead to a substantial reduction in the final trajectory reward, encouraging the agent to maintain safe and efficient memory practices.


%% file: sections/5_experiment.tex
\section{Experiments}

\subsection{Experimental Setup}

\textbf{Datasets.} To comprehensively evaluate AgeMem, we select five widely-used datasets in LLM-based agent research: ALFWorld~\citep{shridhar2020alfworld}, SciWorld~\citep{wang2022scienceworld}, PDDL~\citep{chang2024agentboard}, BabyAI~\citep{chevalier2018babyai}, and HotpotQA~\citep{yang2018hotpotqa}. These datasets cover embodied action, game-based reasoning, and knowledge-intensive question answering, providing diverse evaluation scenarios. Since the HotpotQA dataset contains both questions and supporting facts, automatically providing Stage~1 contextual information, AgeMem is fine-tuned with RL \textit{only on the HotpotQA training set} and then evaluated directly on all datasets. Detailed dataset statistics are provided in Appendix~\ref{app:datasets}.
\\
\textbf{Evaluation metrics.} For the primary task completion metrics, we adopt Success Rate (SR) for ALFWorld, SciWorld, and BabyAI, Progress Rate (PR) for PDDL, and LLM-as-a-Judge (J) for HotpotQA. Additionally, we employ an LLM-based evaluator to assess the quality of stored long-term memory during knowledge reasoning, measured by Memory Quality (MQ). The prompts of the LLM-based evaluation are provided in Appendix~\ref{app:evaluation}.
\\
\textbf{Baselines \& LLM backbones.} We compare AgeMem against four representative agent LTM systems: LangMem~\citep{langmem2025}, A-Mem~\citep{xu2025mem}, Mem0~\citep{chhikara2025mem0}, and $\text{Mem0}^g$ (a graph-based variant officially provided as part of Mem0). To better demonstrate the effectiveness of RL training, we also include AgeMem-noRL, which is not fine-tuned with RL. In ablation studies on STM, we compare STM tools with the RAG approach. For the base agent models, we use Qwen2.5-7B-Instruct and Qwen3-4B-Instruct. More baseline configurations are in Appendix~\ref{app:baselines}.
\\
\textbf{Implementation details.} We build agents using the Agentscope framework~\citep{gao2025agentscope} and fine-tune AgeMem using the Trinity framework~\citep{pan2025trinity}. Further implementation details are provided in Appendix~\ref{app:implementation}.

\begin{table*}[t]
\centering
\small
\caption{Performance comparison across five benchmarks. The \textbf{best} and \underline{second-best} results are marked.}
\label{tab:main_results}
\setlength{\tabcolsep}{6pt}
\begin{tabular}{@{}l|l|@{\hspace{6pt}}ccccc@{\hspace{6pt}}|c@{}}
\toprule
\textbf{LLM Backbone} & \textbf{Method} & \textbf{ALFWorld} & \textbf{SciWorld} & \textbf{PDDL} & \textbf{BabyAI} & \textbf{HotpotQA} & \textbf{Average} \\
\midrule
\multirow{7}{*}{\textbf{Qwen2.5-7B-Instruct}} 
& No-Memory & 27.16 & 13.80 & 10.15 & 50.80 & 38.36 & 28.05 \\
& LangMem & \underline{38.27} & 28.29 & 15.85 & 51.34 & 37.43 & 34.23 \\
& A-Mem & 34.68 & 28.06 & \textbf{18.39} & 58.82 & 43.95 & 36.78 \\
& Mem0 & 37.49 & 26.99 & 13.96 & \underline{60.58} & \underline{46.66} & \underline{37.14} \\
& Mem0$^{g}$ & 35.34 & \underline{30.50} & 14.86 & 58.78 & 42.06 & 36.31 \\
& \cellcolor{gray!15}AgeMem-noRL & \cellcolor{gray!15}37.90 & \cellcolor{gray!15}28.67 & \cellcolor{gray!15}8.87 & \cellcolor{gray!15}46.34 & \cellcolor{gray!15}45.36 & \cellcolor{gray!15}33.43 \\
& \cellcolor{gray!15}AgeMem (Ours) & \cellcolor{gray!15}\textbf{41.07} & \cellcolor{gray!15}\textbf{35.55} & \cellcolor{gray!15}\underline{17.31} & \cellcolor{gray!15}\textbf{61.42} & \cellcolor{gray!15}\textbf{54.44} & \cellcolor{gray!15}\textbf{41.96} \\
\midrule
\multirow{7}{*}{\textbf{Qwen3-4B-Instruct}} 
& No-Memory & 38.51 & 47.89 & 30.14 & 55.83 & 47.48 & 43.97 \\
& LangMem & 40.89 & 50.42 & 28.42 & 53.80 & 42.70 & 43.25 \\
& A-Mem & 34.31 & 50.14 & \underline{34.41} & \underline{61.35} & 48.48 & \underline{45.74} \\
& Mem0 & \underline{41.17} & \underline{51.38} & 31.72 & 60.05 & 39.16 & 44.70\\
& Mem0$^{g}$ & 36.69 & 47.76 & 29.61 & 57.59 & 38.12 & 41.95 \\
& \cellcolor{gray!15}AgeMem-noRL & \cellcolor{gray!15}38.02 & \cellcolor{gray!15}50.42 & \cellcolor{gray!15}27.52 & \cellcolor{gray!15}57.48 & \cellcolor{gray!15}\underline{54.49} & \cellcolor{gray!15}45.59 \\
& \cellcolor{gray!15}AgeMem (Ours) & \cellcolor{gray!15}\textbf{48.97} & \cellcolor{gray!15}\textbf{59.48} & \cellcolor{gray!15}\textbf{35.07} & \cellcolor{gray!15}\textbf{72.56} & \cellcolor{gray!15}\textbf{55.49} & \cellcolor{gray!15}\textbf{54.31} \\
\bottomrule
\end{tabular}
\vspace{-0.3cm}
\end{table*}

\subsection{Main Results}
\textbf{Comparison with counterparts.} Table~\ref{tab:main_results} shows that AgeMem achieves the highest average performance on both Qwen2.5-7B-Instruct (41.96\%) and Qwen3-4B-Instruct (54.31\%), outperforming all baselines across five datasets with relative gains of 49.59\% and 23.52\% over no-memory, respectively. Compared to the best baselines (Mem0 and A-Mem), AgeMem improves by 4.82 and 8.57 percentage points on average. RL training contributes 8.53 and 8.72 percentage points of improvement over AgeMem-noRL, validating the three-stage progressive RL strategy.
\begin{figure}[t]
\centering
\includegraphics[width=0.4\textwidth]{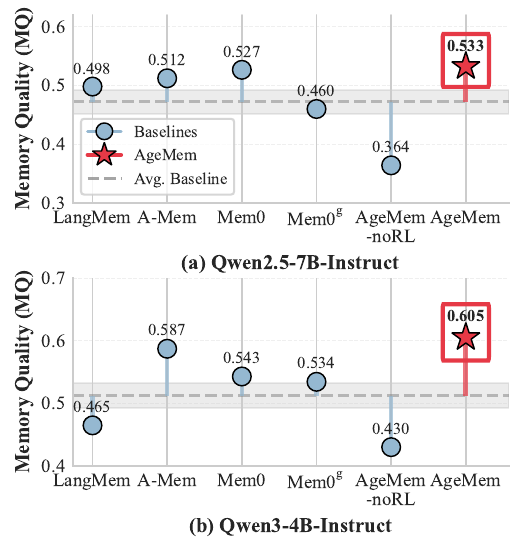}
\caption{Memory Quality scores for different methods on HotpotQA. Higher scores indicate better relevance between stored memories and ground-truth facts.}
\label{fig:memory_quality}
\end{figure}
\begin{table}[t]
\centering
\small
\caption{Tool usage statistics on HotpotQA. Numbers show average calls per episode.}
\label{tab:tool_usage}
\begin{tabular}{@{}l|cc|cc@{}}
\toprule
\multirow{2}{*}{\textbf{Tool Category}} & \multicolumn{2}{c|}{\textbf{Qwen2.5-7B}} & \multicolumn{2}{c}{\textbf{Qwen3-4B}} \\
\cmidrule(lr){2-3} \cmidrule(lr){4-5}
 & noRL & GRPO & noRL & GRPO \\
\midrule
\multicolumn{5}{c}{\textbf{\emph{LTM Tool Statistics}}} \\
\midrule
\textsc{Add} Memory & 0.92 & 1.64 & 2.49 & 2.64 \\
\textsc{Update} Memory & 0.00 & 0.13 & 0.13 & 0.34 \\
\textsc{Delete} Memory & 0.00 & 0.08 & 0.00 & 0.22 \\
\midrule
\multicolumn{5}{c}{\textbf{\emph{STM Tool Statistics}}} \\
\midrule
\textsc{Retrieve} Memory & 2.31 & 1.95 & 4.62 & 4.35 \\
\textsc{Summary} Context & 1.08 & 0.82 & 0.11 & 0.96 \\
\textsc{Filter} Context & 0.02 & 0.31 & 0.15 & 0.16 \\
\midrule
\textbf{Total Calls} & 4.33 & 4.92 & 7.50 & 8.67 \\
\bottomrule
\end{tabular}%
\vspace{-0.3cm}
\end{table}
\begin{figure}[t]
\centering
\includegraphics[width=0.48\textwidth]{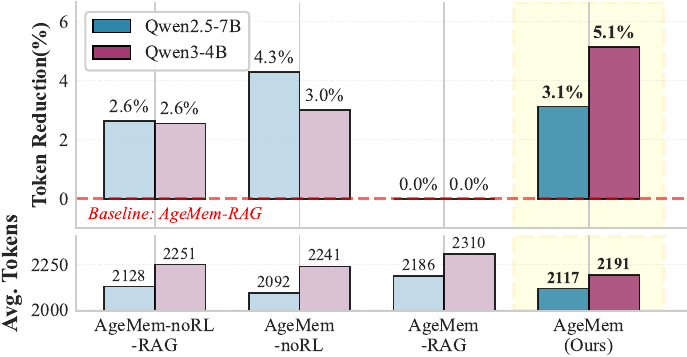}
\caption{Average prompt token counts under different STM management configurations on HotpotQA. The suffix ``-RAG'' indicates the adoption of RAG in place of STM tool-based management.}
\label{fig:stm_management}
\end{figure}
\\
\textbf{Quality of stored long-term memories.} To evaluate the quality of stored memories, we leverage the ground-truth facts provided in the HotpotQA dataset and assess the relevance between stored memories and these facts using an LLM-based evaluator. Figure~\ref{fig:memory_quality} presents the Memory Quality (MQ) scores for different baselines. AgeMem achieves the highest memory quality on both model backbones, with MQ scores of 0.533 and 0.605, respectively. This indicates that the unified memory management framework not only improves task performance but also promotes the storage of high-quality, reusable knowledge. The comparison with baseline methods further validates that AgeMem's tool-based memory operations lead to more selective and higher-quality memory construction.
\\
\textbf{Effectiveness of STM management.} We evaluate the effectiveness of STM management by measuring the prompt token count under different configurations on HotpotQA. Figure~\ref{fig:stm_management} shows that AgeMem successfully reduces prompt token usage compared to variants without STM tools (-RAG). On Qwen2.5-7B-Instruct, AgeMem uses 2,117 tokens on average, compared to 2,186 tokens for AgeMem-RAG, representing a reduction of 3.1\%. On Qwen3-4B-Instruct, the reduction is even more pronounced: AgeMem uses 2,191 tokens versus 2,310 tokens for AgeMem-RAG, a reduction of 5.1\%. These results demonstrate that the learned STM management tools effectively control context expansion, enabling more efficient token usage while maintaining task performance.
\begin{figure*}[t]
  \centering
  \includegraphics[width=0.9\textwidth]{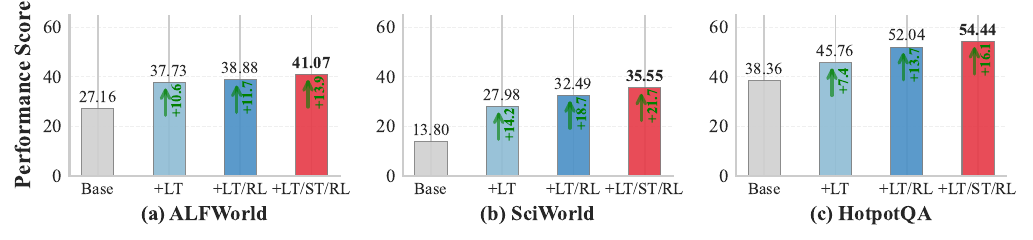}
  \caption{Ablation study on LTM, STM, and RL components (Qwen2.5-7B-Instruct). 
  \textbf{Base}: No-memory baseline; \textbf{+LT}: AgeMem-noRL-RAG (LTM tools only); 
  \textbf{+LT/RL}: AgeMem-RAG (RL with LTM tools); 
  \textbf{+LT/ST/RL}: AgeMem (full AgeMem system with RL). 
  Green arrows indicate performance gains over the baseline. }
  \label{fig:ablation_components}
\end{figure*}
\\
\textbf{Tool usage analysis.}
Table~\ref{tab:tool_usage} reports tool usage statistics before and after RL fine-tuning on HotpotQA. RL training substantially increases the use of long-term memory tools, especially \textsc{Add} and \textsc{Update}. On Qwen2.5-7B-Instruct, \textsc{Add} operations rise from 0.92 to 1.64, and \textsc{Update} operations appear after training (0.13 v.s.\ nearly zero). Similar trends are observed on Qwen3-4B-Instruct, with higher frequencies of both \textsc{Add} and \textsc{Update}.
For short-term memory tools, RL leads to more balanced tool usage. The frequency of \textsc{Filter} increases notably (e.g., from 0.02 to 0.31 on Qwen2.5), indicating proactive context control.

Notably, the decrease in \textsc{Retrieve} frequency after RL training (Qwen2.5: $2.31\!\to\!1.95$; Qwen3: $4.62\!\to\!4.35$) reflects a qualitative shift in retrieval strategy rather than undertraining. Before RL, the agent retrieves reactively and repeatedly to compensate for suboptimal Stage-1 storage. After RL optimization, \textsc{Add}/\textsc{Update} frequencies increase (Table~\ref{tab:tool_usage}), improving LTM quality; retrieval then becomes more selective and query-driven, used primarily when previously stored information is genuinely needed. This reduction in retrieval frequency coincides with improved task performance and MQ, indicating greater efficiency rather than insufficient learning. Overall, these patterns suggest that RL training enables coordinated and adaptive memory management. Detailed case studies are provided in Appendix~\ref{app:case_study}.

\begin{figure}[t]
\centering
\includegraphics[width=0.8\columnwidth]{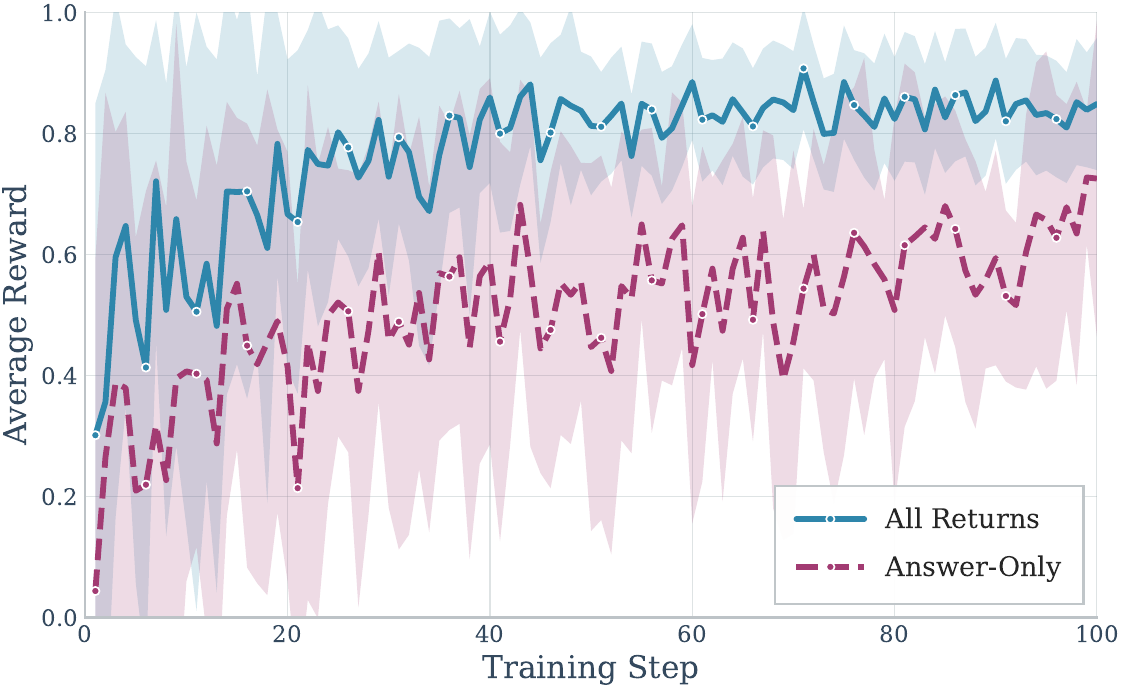}
\caption{Training convergence curves on Qwen2.5-7B-Instruct comparing All-Returns (solid line) v.s. Answer-Only (dashed line) reward strategies.}
\label{fig:reward_convergence}
\vspace{-4mm}
\end{figure}

\subsection{Ablation Studies}
\textbf{LTM-STM components.}
To validate the contributions of individual components, we conduct ablation studies on LTM, STM, and RL training. Figure~\ref{fig:ablation_components} presents results on three representative datasets using Qwen2.5-7B-Instruct as the backbone (results for Qwen3-4B-Instruct are provided in Appendix~\ref{app:ablation}).
Adding LTM alone (+LT) yields substantial gains of +10.6\%, +14.2\%, and +7.4\% over the baseline. Incorporating RL training (+LT/RL) further improves performance, particularly on HotpotQA (+6.3\%), demonstrating the effectiveness of our reward-based optimization. The full AgeMem system (+LT/ST/RL) achieves the best results across all benchmarks, with overall improvements of +13.9\%, +21.7\%, and +16.1\%. Notably, adding STM tools provides the most significant boost on SciWorld (+3.1\%) and HotpotQA (+2.4\%), validating that learned context management outperforms static RAG approaches. These progressive improvements confirm that unified memory management with end-to-end RL is essential for optimal agent performance.
\\
\textbf{Reward function.}
To demonstrate the effectiveness of our multi-component reward function design, we compare the full reward function (All-Returns) against a variant using only $R_{\text{task}}$ (Answer-Only).
Figure~\ref{fig:reward_convergence} shows the reward convergence curves of Qwen2.5-7B-Instruct during GRPO training on HotpotQA. The full reward function leads to significantly faster convergence and higher final performance compared to the task-only variant. As detailed in Table~\ref{tab:reward_ablation}, the All-Returns strategy achieves higher LLM-as-a-Judge scores (0.544 v.s. 0.509) while maintaining substantially better memory quality (0.533 v.s. 0.479).  Notably, despite using more tokens (2117 v.s. 2078), the All-Returns strategy achieves better overall performance, indicating that the additional context and memory operations contribute meaningfully to reasoning quality. Similar patterns are observed on Qwen3-4B-Instruct (see Appendix~\ref{app:qwen3_results}).
\\
\textbf{FILTER threshold $\theta_{f}$.}
Table~\ref{tab:theta} reports performance on HotpotQA as $\theta_{f}$ varies. Performance is stable within $\theta_{f} \in [0.4, 0.8]$, indicating that AgeMem is not highly sensitive to precise threshold tuning. When $\theta_{f}$ is too low, filtering becomes too 
aggressive, discarding potentially useful context and increasing reliance on later reconstruction. When $\theta_{f}$ is too high, filtering becomes overly permissive, allowing marginal context to pass through STM and slightly degrading memory quality. Average token counts remain similar across settings, confirming that differences stem from selection quality rather than context length alone.

\begin{table}[t]
\centering
\small
\caption{Reward function ablation on HotpotQA using Qwen2.5-7B-Instruct. All-Returns v.s. Answer-Only reward strategies. ``TN'' is the token number, and ``TC'' denotes the number of tool calls.}
\label{tab:reward_ablation}
\begin{tabular}{@{}lcccc@{}}
\toprule
\textbf{Strategy} & \textbf{J}($\uparrow$) & \textbf{TN}($\downarrow$) & \textbf{MQ}($\uparrow$) & \textbf{TC}(-) \\
\midrule
Answer-Only & 0.509 & \textbf{2078} & 0.479 & 3.93 \\
All-Returns & \textbf{0.544} & 2117 & \textbf{0.533} & 4.92 \\
\bottomrule
\end{tabular}
\vspace{-0.2cm}
\end{table}
\begin{table}[t]
\centering
\small
\caption{Sensitivity of AgeMem to the FILTER threshold on 
HotpotQA.}
\label{tab:theta}
\begin{tabular}{ccccc}
\toprule
$\theta_{f}$ & \textbf{J}($\uparrow$) & \textbf{MQ}($\uparrow$) & \textbf{Avg.\ Tokens} \\
\midrule
0.4 & 0.524 & 0.511 & 2089 \\
0.5 & 0.551 & 0.550 & 2116 \\
0.6 & 0.544 & 0.533 & 2117 \\
0.7 & 0.530 & 0.526 & 2149 \\
0.8 & 0.531 & 0.510 & 2134 \\
\bottomrule
\end{tabular}
\vspace{-5mm}
\end{table}

%% file: sections/6_conclusion.tex
\section{Conclusion}
In this work, we propose Agentic Memory (AgeMem), a unified memory management framework that enables LLM-based agents to jointly control long-term and short-term memory through learnable, tool-based actions. By integrating memory operations directly into the agent’s policy and training them with a progressive reinforcement learning strategy, AgeMem replaces heuristic memory pipelines with an end-to-end optimized solution. Extensive experiments across diverse long-horizon benchmarks show that AgeMem improves both task performance and memory quality while maintaining efficient context usage. These results highlight the importance of unified, agent-centric memory policies and suggest a promising direction for building scalable and adaptive LLM agents capable of long-term reasoning.

%% file: sections/appendix.tex
\appendix
\section{Detailed Design and Implementation of AgeMem}
\label{app:details}

This appendix provides full technical details omitted from the main text due to space constraints. We first present precise definitions and pseudo-formulations for each memory-management tool (Appendix~\ref{app:tools}), then give implementable formulas for the reward components used in training (Appendix~\ref{app:rewards}). Finally, we provide the complete algorithmic specification (Appendix~\ref{app:algorithm}).

\subsection{Memory Management Tools}
\label{app:tools}

AgeMem exposes a small set of structured tools that the agent may invoke as part of its action \(a_t\). Each tool is implemented as a deterministic or stochastic function that transforms the short-term context \(C_t\), the long-term memory store \(\mathcal{M}_t\), or both. Unlike traditional memory systems that rely on external heuristics or predefined schedules, AgeMem integrates these tools directly into the agent's action space, enabling the model to learn when and how to use each tool through reinforcement learning. Below we give precise operational definitions, implementation details, and the system prompts that guide tool usage.

\paragraph{Notation.}  
Long-term memory store at time \(t\) is \(\mathcal{M}_t = \{m_i\}_{i=1}^{|\mathcal{M}_t|}\), where each memory \(m_i\) contains a content string and optional metadata. Short-term context is \(C_t = [u_1, u_2, \ldots, u_{n_t}]\) (message list), and \(\text{enc}(\cdot)\) denotes a text encoder that returns a dense embedding. We use cosine similarity for semantic matching throughout the framework.

\paragraph{\textsc{Retrieve}.}
The \textsc{Retrieve} operation enables the agent to access relevant information from long-term memory based on semantic similarity. This operation is crucial for bringing stored knowledge into the active context when needed for reasoning. The retrieval operation returns the top-$k$ most similar memories to the query $q$:
\begin{equation}
\textsc{Retrieve}(q, k) = \text{TopK}(\mathcal{M}_t, \; \text{sim}(q, m_i), \; k),
\end{equation}
where the similarity function is defined as:
\begin{equation}
\text{sim}(q, m_i) =
\frac{\text{enc}(q)^\top \, \text{enc}(m_i)}
{\|\text{enc}(q)\| \, \|\text{enc}(m_i)\|}.
\end{equation}
The retrieved memories are then inserted into the short-term context $C_t$, making them available for immediate reasoning. The parameter $k$ controls the number of memories retrieved, typically set to 3-5 in our experiments to balance relevance and context size.

\paragraph{\textsc{Add}.}
The \textsc{Add} operation allows the agent to store new information in long-term memory for future use. This operation is essential for accumulating knowledge across interactions and sessions. A new memory entry is created by:
\begin{equation}
m_{\text{new}} = \big(c, \text{enc}(c), \text{metadata}\big),
\end{equation}
where $c$ is the content to be stored, $\text{enc}(c)$ is its embedding vector, and $\text{metadata}$ includes timestamp, source information, and optional tags. The memory store is then updated:
\begin{equation}
\mathcal{M}_{t+1} = \mathcal{M}_t \cup \{m_{\text{new}}\}.
\end{equation}
The agent learns to identify salient information worth storing through the reward function, which encourages storing high-quality, reusable knowledge while penalizing redundant or irrelevant entries.

\paragraph{\textsc{Update} and \textsc{Delete}.}

Memory maintenance operations enable the agent to keep its long-term memory store current and relevant. The \textsc{Update} operation modifies existing memories when new information supersedes or refines previous knowledge. For an existing memory $m_i$, the update operation is defined as:
\begin{equation}
m_i \leftarrow \big(c^{\prime}, \text{enc}(c^{\prime}), \text{metadata}^{\prime}\big),
\end{equation}
where $c^{\prime}$ is the updated content and $\text{metadata}^{\prime}$ reflects the modification timestamp. The \textsc{Delete} operation removes obsolete or incorrect memories:
\begin{equation}
\mathcal{M}_{t+1} = \mathcal{M}_t \setminus \{m_i\}.
\end{equation}
These operations are particularly important in long-horizon tasks where information may become outdated or where the agent needs to correct earlier mistakes. The reward function encourages meaningful updates and deletions that improve memory quality over time.

\paragraph{\textsc{Summary}.}
The \textsc{Summary} operation compresses conversation history in the short-term context to prevent context overflow while preserving essential information. This operation is critical for managing long conversations that exceed context window limits. Given a subset of context indices $s$, the summary operation is defined as:
\begin{equation}
C_t^{\prime}
= C_t \setminus \{u_i \mid i \in s\}
\, \cup \, \{\text{Summarize}(\{u_i\}_{i \in s})\},
\end{equation}
where $\text{Summarize}(\cdot)$ is implemented by LLM with a summarization system prompt. The agent can specify which messages to summarize using the \texttt{`span'} parameter, which can be:
\begin{itemize}
\item \texttt{``all''}: Summarize all non-system messages.
\item \texttt{``N''}: Summarize the last $N$ messages.
\end{itemize}
The summarization process uses the following system prompt to ensure high-quality compression:

\begin{lstlisting}[
    language=,
    basicstyle=\footnotesize\ttfamily,
    breaklines=true,
    columns=flexible,
    keepspaces=true,
    frame=single,
    frameround=tttt,
    framesep=5pt
]
You are a conversation summarization assistant. 
Your goal is to compress the given conversation span into a concise summary that preserves all important information, intentions, decisions, and unresolved questions. 
The summary will later be used to replace the original conversation in the context, so make sure nothing essential is lost.

Instructions:
1. Read the provided conversation rounds carefully.
2. Identify the main topics, actions, results, and open issues.
3. Write a clear, factual summary in natural language.
4. Do NOT include greetings, filler text, or redundant phrasing.

Input:
- Conversation content: [CONVERSATION_TEXT]

Output:
- A concise yet comprehensive summary of the above conversation span.

Let's start the conversation summarization.
\end{lstlisting}
The agent learns to invoke summarization proactively before context overflow occurs, balancing information preservation with efficiency.

\paragraph{\textsc{Filter}.}
The \textsc{Filter} operation filters out irrelevant or redundant messages from the short-term context based on semantic similarity. This operation helps maintain a focused context by filtering out noise and distractions. Specifically, it removes messages whose similarity to a given criteria $c$ exceeds a threshold $\theta_{f}$:
\begin{equation}
C_t^{\prime}
= \left\{ u_i \in C_t
\, \middle| \,
\text{sim}(c, u_i) < \theta_{f}
\right\}.
\end{equation}
In all experiments, we set $\theta_{f} = 0.6$ by default. The criteria $c$ can be specified by the agent (e.g., a description of what to keep) or can be automatically derived from the current task context. This operation is particularly useful in Stage~2 of training, where distractors are introduced to test the agent's ability to filter irrelevant information.

\paragraph{Tool invocation as structured actions.}  
Each tool is exposed via a schema specifying its function name and required arguments. The agent's policy outputs either language tokens (for text generation) or structured tool calls (for memory operations). The agent is guided by a system prompt that defines the tool-calling interface and response format. The system prompt used in AgeMem is as follows:

\begin{lstlisting}[
    language=,
    basicstyle=\footnotesize\ttfamily,
    breaklines=true,
    columns=flexible,
    keepspaces=true,
    frame=single,
    frameround=tttt,
    framesep=5pt
]
You are an intelligent assistant that solves complex problems by managing context and memory with tools when needed.

## Available Tools:[TOOLS]

## Problem-Solving Workflow
You must follow a structured reasoning and action process for every task:
1. **Think & Plan**  
   Always start with a <think>...</think> block.  
   Inside it, explain your reasoning, plan your next step, and decide whether you need to call a tool or provide a final answer.
2. **Tool Calls**  
   If you decide to use one or more tools, follow your <think> block with a <tool_call>...</tool_call> block.  
   - You may call **one or multiple tools** in a single step.  
   - List multiple tool calls as elements of a JSON array.  
   - Each tool call must include "name" and "arguments".  
   - Example:
     <tool_call>[{{"name": "Retrieve_memory", "arguments": {{"query": "math problem solving strategies", "top_k": 3}}}}, {{"name": "Add_memory", "arguments": {{"content": "Strategy summary for reuse", "memory_type": "problem_solving"}}}}]</tool_call>
3. **Final Answer**  
   When you no longer need tools and are ready to present your final output, follow your last <think> block with an <answer>...</answer> block containing the full response.
4. **Mutual Exclusivity Rule**  
   After **each <think> block**, you must choose exactly **one** of the following:
   - a "<tool_call>" block (if you need tools), **or**
   - an "<answer>" block (if you are ready to respond).  
   You must **never** include both "<tool_call>" and "<answer>" immediately after the same "<think>" block.
5. **Iterative Solving**  
   You may repeat this sequence as needed:  
   "<think>" -> "<tool_call>" -> "<think>" -> "<tool_call>" ... -> "<think>" -> "<answer>"  
   until the problem is completely solved.

## Response Format (Strict)
Your full output must follow these rules:
- Every reasoning step must appear inside <think> tags.  
- Every tool usage must appear inside one <tool_call> tag (even if it includes multiple tool invocations).  
- The final solution must be wrapped in <answer> tags.  
- No text should appear outside these tags.

## Guidelines
- Always start with reasoning (<think>).
- After each reasoning step, decide: call tool(s) or answer.
- You can call multiple tools within one <tool_call> JSON array.
- Be concise, logical, and explicit in reasoning.
- Manage memory actively: retrieve, add, update, summarize, filter, or delete as needed.
- Use <answer> only once when the final solution is ready.

Let's start!
\end{lstlisting}
This prompt structure ensures that the agent follows a consistent format for reasoning, tool invocation, and final answers, which is essential for reliable parsing and reward computation during RL training. The structured format also enables the agent to coordinate multiple memory operations within a single reasoning step, supporting efficient unified memory management.

Figure~\ref{fig:stm_tool_schemas} and~\ref{fig:ltm_tool_schemas} present our tool schemas for short-term memory and long-term memory management, showing the exact function signatures and argument types that the agent can invoke.

\begin{figure*}[p]
\centering
\begin{tcolorbox}[
    enhanced,
    colback=blue!3!white,
    colframe=blue!40!black,
    fonttitle=\bfseries\sffamily,
    title={\small Short-term Memory (STM) Management Tools},
    coltitle=white,
    attach boxed title to top left={yshift=-2mm, xshift=3mm},
    boxed title style={colback=blue!40!black, sharp corners},
    sharp corners,
    boxrule=0.5pt,
    left=3pt,
    right=3pt,
    top=8pt,
    bottom=3pt,
    arc=0pt
]
\begin{lstlisting}[
    language=Python,
    basicstyle=\footnotesize\ttfamily,
    keywordstyle=\color{blue!70!black}\bfseries,
    stringstyle=\color{red!60!black},
    commentstyle=\color{green!50!black}\itshape,
    numberstyle=\tiny\color{gray},
    showstringspaces=false,
    breaklines=true,
    columns=flexible,
    keepspaces=true,
    tabsize=4,
    xleftmargin=0pt,
    xrightmargin=0pt,
    frame=none
]
STM_TOOLS = [
    {
        "name": "Summary_context",
        "description": "Summarizes conversation rounds to reduce tokens while preserving key information.",
        "parameters": {
            "properties": {
                "span": {
                    "description": "The range of conversation rounds to summarize. Can be 'all' for entire context, or a number (e.g., '5') for the last N rounds. A system, user, assistant and 'tool' message are considered as one round.",
                    "type": "string"
                }
            },
            "required": ["span"]
        }
    },
    {
        "name": "Filter_context",
        "description": "Filters out irrelevant or outdated content from the conversation context to improve task-solving efficiency. ",
        "parameters": {
            "properties": {
                "criteria": {
                    "description": "The criteria for content removal. Can be keywords, phrases, or descriptions of content types to remove (e.g., 'the birthday of John', 'the age of Mary').",
                    "type": "string"
                }
            },
            "required": ["criteria"]
        }
    },
    {
        "name": "Retrieve_memory",
        "description": "Retrieves relevant memories and adds them to current context.",
        "parameters": {
            "properties": {
                "query": {
                    "description": "The search query to find relevant memories. Should describe what kind of information or context is needed.",
                    "type": "string"
                },
                "top_k": {
                    "description": "The maximum number of memories to retrieve. Defaults to 3.",
                    "type": "integer"
                },
                "metadata_filter": {
                    "description": "Optional metadata filters to narrow down memory search (e.g., {'type': 'user_info', 'domain': 'math'}).",
                    "type": "object"
                }
            },
            "required": ["query"]
        }
    }
]
\end{lstlisting}
\end{tcolorbox}
\caption{Short-term memory (STM) management tools for conversational context management. These tools enable summarization, selective filtering, and retrieval operations to maintain efficient context windows.}
\label{fig:stm_tool_schemas}
\end{figure*}

\begin{figure*}[p]
\centering
\begin{tcolorbox}[
    enhanced,
    colback=green!3!white,
    colframe=green!40!black,
    fonttitle=\bfseries\sffamily,
    title={\small Long-term Memory (LTM) Management Tools},
    coltitle=white,
    attach boxed title to top left={yshift=-2mm, xshift=3mm},
    boxed title style={colback=green!40!black, sharp corners},
    sharp corners,
    boxrule=0.5pt,
    left=3pt,
    right=3pt,
    top=8pt,
    bottom=3pt,
    arc=0pt
]
\begin{lstlisting}[
    language=Python,
    basicstyle=\footnotesize\ttfamily,
    keywordstyle=\color{blue!70!black}\bfseries,
    stringstyle=\color{red!60!black},
    commentstyle=\color{green!50!black}\itshape,
    numberstyle=\tiny\color{gray},
    showstringspaces=false,
    breaklines=true,
    columns=flexible,
    keepspaces=true,
    tabsize=4,
    xleftmargin=0pt,
    xrightmargin=0pt,
    frame=none
]
LTM_TOOLS = [
    {
        "name": "Add_memory",
        "description": "Adds new information to external memory store for future reference.",
        "parameters": {
            "properties": {
                "content": {
                    "description": "The content to store in memory.",
                    "type": "string"
                },
                "metadata": {
                    "description": "Optional metadata tags to categorize and filter the memory.",
                    "type": "object"
                },
                "memory_type": {
                    "description": "The type of memory being stored.",
                    "type": "string"
                }
            },
            "required": ["content"]
        }
    },
    {
        "name": "Update_memory",
        "description": "Updates existing memory. Requires memory_id from prior retrieval.",
        "parameters": {
            "properties": {
                "memory_id": {
                    "description": "The unique identifier of the memory to update. Must be obtained from a previous memory retrieval operation.",
                    "type": "string"
                },
                "content": {
                    "description": "The new content to replace the existing memory content.",
                    "type": "string"
                },
                "metadata": {
                    "description": "Updated metadata for the memory.",
                    "type": "object"
                }
            },
            "required": ["memory_id", "content"]
        }
    },
    {
        "name": "Delete_memory",
        "description": "Removes memory from store. Requires confirmation.",
        "parameters": {
            "properties": {
                "memory_id": {
                    "description": "The unique identifier of the memory to delete. Must be obtained from a previous memory retrieval operation.",
                    "type": "string"
                },
                "confirmation": {
                    "description": "Confirmation that this memory should be permanently deleted.",
                    "type": "boolean"
                }
            },
            "required": ["memory_id", "confirmation"]
        }
    }
]
\end{lstlisting}
\end{tcolorbox}
\caption{Long-term memory (LTM) management tools for persistent storage. These tools provide add, update, and delete capabilities for maintaining long-term information retention across conversations.}
\label{fig:ltm_tool_schemas}
\end{figure*}

\subsection{Reward Function Design}
\label{app:rewards}

This section provides implementable formulas for the reward components described in the main text. All component scores are normalized to \([0,1]\) (unless noted) to enable stable weighting.
\\
\textbf{Overview.} The overall trajectory-level reward is defined as:
\begin{equation}
R(\tau) = \mathbf{w}^\top \mathbf{R} + P_{\text{penalty}},
\end{equation}
where $\mathbf{w} = [w_{\text{task}}, w_{\text{context}}, w_{\text{memory}}]^\top$ are tunable weights, $\mathbf{R} = [R_{\text{task}}, R_{\text{context}}, R_{\text{memory}}]^\top$ denote task completion, context management, and memory management rewards respectively, and $P_{\text{penalty}}$ penalizes undesired behaviors.
\\
\textbf{Task completion reward \(R_{\text{task}}\).}
Let the agent produce a final answer \(A_{\text{pred}}\). We obtain a judge score \(S_{\text{judge}}(A_{\text{pred}}, A_{q})\in[0,1]\) via an evaluator (LLM judge), where $A_{q}$ denotes the expected ground truth. Then the task reward $R_{\text{task}}$ is:
\begin{equation}
R_{\text{task}}
=
\begin{cases}
S_{\text{judge}}(A_{\text{pred}}, A_{q}),
& \text{if has answer}, \\
P_{\text{no-answer}},
& \text{otherwise},
\end{cases}
\end{equation}
with \(P_{\text{no\_answer}}=-1.0\) by default.
\\
\textbf{Context management reward \(R_{\text{context}}\).}
We decompose the overall context management reward into three normalized components that jointly evaluate how effectively the model maintains a compact yet information-preserving context state. Formally, we define: 
\begin{equation}
R_{\text{context}} = \sum_{i=1}^{3} \alpha_i R_i,
\end{equation}
where $R_i \in \{R_{\text{compression}}, R_{\text{preventive}}, R_{\text{preservation}}\}$, $\sum_i \alpha_i = 1$, and we use uniform weights $\alpha_i = 1/3$ unless otherwise specified. For \textbf{compression efficiency,} we evaluate the compactness of the final context $C_t$ by computing
\begin{equation}
R_{\text{compression}}
= \max\!\left(0,\; 1 - \frac{T_{\text{used}}}{T_{\text{max}}}\right),
\end{equation}
where $T_{\text{used}}$ denotes the number of tokens present in the context when the final answer is generated, and $T_{\text{max}}$ is the allowed budget.
For \textbf{preventive management}, we define $R_\text{preventive}$ to assess proactive behavior:
\begin{equation}
R_{\text{preventive}} = \mathbbm{1}[\text{tool invoked before overflow}],
\end{equation}
which equals 1 when the model invokes a context-reduction tool before reaching the token limit, and 0 otherwise.
For \textbf{information preservation}, we identify a set of key tokens or phrases $K_q$ extracted from the user query $q$, such as named entities or temporal and spatial expressions. Let $\mathbbm{1}_{\text{preserve}}$ indicate whether these items remain present (either directly or via a retained summary) at the time of answer generation. The preservation reward is therefore
\begin{equation}
R_{\text{preservation}}
= \mathbbm{1}_{\text{preserve}}.
\end{equation}
\\
\textbf{Memory management reward $R_{\text{memory}}$.}
The memory management reward consists of three key components that evaluate storage quality, maintenance operations, and semantic relevance. We define it as:
\begin{equation}
R_{\text{memory}} = \sum_{j=1}^{3} \beta_j R_j,
\end{equation}
where $R_j \in \{R_{\text{storage}}, R_{\text{maintenance}}, R_{\text{relevance}}\}$, $\sum_j \beta_j = 1$, and we use uniform weights $\beta_j = 1/3$ unless otherwise specified.
For \textbf{Storage Quality}, during the memory storage process in Stage~1, the agent may add $N_{\text{total}}$ memory entries, among which $N_{\text{high\_quality}}$ are identified as high-quality based on an LLM’s analysis of the input query $q$ and its expected answer $A_q$. The storage quality reward is defined as the proportion of high-quality memories:
\begin{equation}
R_{\text{storage}} = \frac{N_{\text{high\_quality}}}{\max(1, N_{\text{total}})}.
\end{equation}
This metric incentivizes the agent to store valuable information while avoiding the accumulation of redundant or low-quality memories.
For \textbf{Maintenance}, to encourage the agent to actively maintain the memory bank, we reward update or delete operations:
\begin{equation}
R_{\text{maintenance}} = \mathbbm{1}[\text{update or delete performed}].
\end{equation}
This mechanism promotes dynamic memory management and timely cleanup.
For \textbf{Semantic Relevance}, to quantify the semantic match between retrieved memories and the query, we introduce an LLM-based relevance assessment. Let $S_{\text{LLM}}(\mathcal{R}, q)$ be the semantic relevance score of the retrieved memory set $\mathcal{R}$ with respect to query $q$, normalized to the interval $[0, 1]$. The semantic relevance reward is defined as:
\begin{equation}
R_{\text{relevance}} = S_{\text{LLM}}(\mathcal{R}, q).
\end{equation}
This component ensures that retrieved memories are semantically aligned with the current task, enhancing overall reasoning quality.
\\
\textbf{Penalty terms \(P_{\text{penalty}}\).}
We penalize major constraint violations to ensure the agent operates within specified limits:
\begin{equation}
P_{\text{penalty}} = \sum_{k=
1}^{2} P_k \cdot \mathbbm{1}[\text{violation}_k],
\end{equation}
where $P_k \in \{P_{\text{rounds}}, P_{\text{overflow}}\}$ and $\text{violation}_k \in \{\mathbbm{1}[N_{\text{rounds}} > N_{\max}], \mathbbm{1}[T_{\text{used}} > T_{\max}]\}$. Here, $N_{\text{rounds}}$ denotes the number of interaction rounds, $N_{\max}$ is the maximum allowed rounds, $T_{\text{used}}$ represents the total token usage, and $T_{\text{max}}$ is the token budget limit. The penalty coefficients are set to $P_{\text{rounds}} = -1$ and $P_{\text{overflow}} = -0.5$ by default.

\subsection{AgeMem Algorithm}
\label{app:algorithm}

This section provides the complete algorithmic specification of \textbf{AgeMem}, our unified memory management framework for LLM-based agents. The training procedure integrates three progressive stages (long-term memory construction, short-term context management under distractors, and integrated task execution) into a single end-to-end reinforcement learning loop. We present the main training algorithm using a two-column layout for compactness (Algorithm~\ref{alg:main-training}--\ref{alg:main-training-2}), followed by detailed rollout procedures for each stage (Algorithms~\ref{alg:stage1-rollout}--\ref{alg:stage3-rollout}).

\paragraph{Training overview (Algorithm~\ref{alg:main-training}--\ref{alg:main-training-2}).}
The core training loop follows a generate-then-optimize paradigm. For each task $q$ in a training batch $\mathcal{B}$, we generate $K$ independent rollout trajectories $\{\tau_k^{(q)}\}_{k=1}^K$ using the current policy $\pi_\theta$. Each trajectory $\tau_k^{(q)} = (\tau_k^{(1)}, \tau_k^{(2)}, \tau_k^{(3)})$ concatenates experiences from all three stages, forming a complete episode from initial memory construction to final task completion. The agent first builds long-term memory from contextual information $I_q$ (Algorithms~\ref{alg:stage1-rollout}), then learns to filter out distracting information while maintaining useful context (Algorithms~\ref{alg:stage2-rollout}), and finally retrieves stored knowledge to finish the target task (Algorithms~\ref{alg:stage3-rollout}). All experiences are collected into a unified buffer $\mathcal{E}$ spanning multiple tasks and rollouts. 

\begin{figure*}[!hbtp]
\begin{minipage}[t]{0.48\textwidth}
\begin{algorithm}[H]
\caption{AgeMem Training (Part 1)}
\label{alg:main-training}
\small
\begin{algorithmic}[1]
\Require Policy $\pi_{\theta}$, reference $\pi_{\text{ref}}$, batch $\mathcal{B}$, rollouts $K$
\Ensure Trained policy $\pi_{\theta^{*}}$
\State Initialize $\theta$ and $\theta_{\text{old}} \gets \theta$
\For{each training iteration}
    \State $\mathcal{E} \gets \emptyset$ \textcolor{gray}{// Init experience buffer}
    \State \textcolor{blue}{// \textbf{Rollout Phase}}
    \For{each task $q \in \mathcal{B}$}
        \State Get context $I_q$ for task $q$
        \State $M_{\text{dis}} \gets \textsc{DistractorGen}(q)$ 
        \For{$k = 1$ to $K$}
            \State $\mathcal{M} \gets \emptyset$ \textcolor{gray}{// Init LTM}
            \State $\tau_k^{(1)} \gets \textsc{Stage1}(I_q, \pi_{\theta}, \theta_{\text{old}}, \mathcal{M})$ 
            \State $C \gets \emptyset$ \textcolor{gray}{// Reset STM}
            \State $\tau_k^{(2)} \gets \textsc{Stage2}(M_{\text{dis}}, \pi_{\theta}, \theta_{\text{old}}, \mathcal{M})$ 
            \State $\tau_k^{(3)} \gets \textsc{Stage3}(q, \pi_{\theta}, \theta_{\text{old}}, \mathcal{M})$ 
            \State $\tau_k^{(q)} \gets \tau_k^{(1)} \oplus \tau_k^{(2)} \oplus \tau_k^{(3)}$ 
            \State $\mathcal{E} \gets \mathcal{E} \cup \tau_k^{(q)}$ 
        \EndFor
    \EndFor
\EndFor
\end{algorithmic}
\end{algorithm}
\end{minipage}
\hfill
\begin{minipage}[t]{0.48\textwidth}
\begin{algorithm}[H]
\caption{AgeMem Training (Part 2)}
\label{alg:main-training-2}
\small
\begin{algorithmic}[1]
\setcounter{ALG@line}{20}
\State \textcolor{blue}{// \textbf{Advantage Computation}}
\For{each group $G_q = \{\tau_k^{(q)}\}_{k=1}^{K}$}
    \State Extract rewards: $\{r_T^{(k,q)}\}_{k=1}^{K}$
    \State $\mu_{G_q} \gets \frac{1}{K}\sum_{k=1}^{K} r_T^{(k,q)}$
    \State $\sigma_{G_q} \gets \sqrt{\frac{1}{K-1}\sum_{k=1}^{K}(r_T^{(k,q)} - \mu_{G_q})^2}$
    \For{each trajectory $\tau_k^{(q)} = (e_1, \ldots, e_T)$}
        \State $A_T^{(k,q)} \gets \frac{r_T^{(k,q)} - \mu_{G_q}}{\sigma_{G_q} + \epsilon}$ 
        \For{$t = 1$ to $T$}
            \State $A_t^{(k,q)} \gets A_T^{(k,q)}$ \textcolor{gray}{// Broadcast}
        \EndFor
    \EndFor
\EndFor
\State \textcolor{blue}{// \textbf{Policy Update}}
\State $J(\theta) \gets \mathbb{E}_{(e_t, A_t) \sim \mathcal{E}} [\rho_t A_t - \beta D_{\text{KL}}[\pi_{\theta}\|\pi_{\text{ref}}]]$
\State $\theta \gets \theta + \eta \nabla_{\theta} J(\theta)$
\State $\theta_{\text{old}} \gets \theta$ 
\State \Return $\pi_{\theta}$
\end{algorithmic}
\end{algorithm}
\end{minipage}
\caption{Main training procedure of AgeMem. For clarity, we split the algorithm into two parts: the rollout phase (left) and the advantage computation with policy update (right).}
\label{fig:main-algo}
\end{figure*}

After the rollout phase, we apply group-based advantage normalization to enable fair comparison across tasks with different reward scales. For each task group $G_q$, terminal rewards $\{r_T^{(k,q)}\}_{k=1}^K$ are normalized to zero mean and unit variance, yielding advantages $A_T^{(k,q)}$ that reflect relative performance within the group. These terminal advantages are then broadcast uniformly to all timesteps within the same trajectory, establishing a consistent learning signal that connects early-stage memory decisions to final task outcomes. This step-wise GRPO mechanism enables long-range credit assignment across heterogeneous operations. The policy is then updated via gradient ascent on the expected advantage, regularized by a KL divergence term to maintain proximity to a reference policy $\pi_{\text{ref}}$ for training stability.

\paragraph{Stage-specific rollout procedures (Algorithm~\ref{alg:stage1-rollout}--\ref{alg:stage3-rollout}).}
The three-stage rollout design reflects the natural progression of memory-augmented task solving. Algorithm~\ref{alg:stage1-rollout} implements the first stage, where the agent engages in casual conversation while being gradually exposed to the contextual information $I_q$. During these $T_1$ exploratory turns, the agent must identify salient information and determine when and which long-term memory tools to invoke—including \textsc{Add}, \textsc{Update}, \textsc{Delete}—to construct an initial memory store $\mathcal{M}$. To support informed memory decisions, the agent proactively performs memory retrieval at every step. This retrieval is not task-driven but serves as an introspective operation: it enables the agent to maintain awareness of the current LTM contents, facilitating decisions about updating or discarding stale entries and ensuring that newly stored information remains coherent with existing knowledge. Since the task query has not yet been revealed in Stage~1, the agent must rely on general cues about which information may become useful later. This encourages the formation of reusable, well-structured memory traces rather than query-specific shortcuts, laying the foundation for effective long-horizon memory management in later stages.

Algorithm~\ref{alg:stage2-rollout} describes the second stage, which deliberately stresses the agent's context management capabilities. The short-term context $C$ is reset to avoid information leakage and facilitate the learning of STM management, while the constructed long-term memory $\mathcal{M}$ persists from Stage 1. Over $T_2$ turns, the agent receives semantically related but ultimately irrelevant distractor messages that could mislead downstream reasoning if left unmanaged. The agent must learn to proactively invoke \textsc{Filter} to filter out low-relevance content based on semantic similarity thresholds, or \textsc{Summary} to compress accumulated context when token budgets become constrained. This stage trains robust filtering strategies that generalize beyond simple heuristics, as the agent receives learning signals from the eventual task performance in Stage~3.



\begin{algorithm}[t]
\caption{Stage 1: LTM Construction}
\label{alg:stage1-rollout}
\begin{algorithmic}[1]
\Require Contextual information $I_q$, policy $\pi_{\theta}$, old params $\theta_{\text{old}}$, memory $\mathcal{M}$, max turn number $N_{max}$
\Ensure Stage 1 trajectory $\tau^{(1)} = (e_1^{(1)}, \ldots, e_{T_1}^{(1)})$
\State Initialize $\tau^{(1)} \gets \emptyset$ and $C \gets \emptyset$
\For{$t = 1$ to $N_{max}$} 
    \State Sample message $m_t \sim I_q$
    \State $\mathcal{M}_{\text{ret}} \gets \textsc{Retrieve}(\mathcal{M}, m_t, k) \cup m_t$ 
    \State $C \gets C \cup \mathcal{M}_{\text{ret}}$ 
    \State $s_t \gets (C, \mathcal{M}, q)$
    \State $a_t \sim \pi_{\theta}(\cdot \mid s_t)$
    
    \State Update $C$ with response from $a_t$
    \State $e_t^{(1)} \gets (s_t, a_t, 0, \log \pi_{\theta_{\text{old}}}(a_t \mid s_t))$
    \State $\tau^{(1)} \gets \tau^{(1)} \cup \{e_t^{(1)}\}$

    \State Memory tool calls from $a_t$ \textcolor{gray}{// Memory Management}
    \If{Output Answer from $a_t$}
        \State Conversation Break
    \EndIf
\EndFor
\State \Return $\tau^{(1)}$
\end{algorithmic}
\end{algorithm}

\begin{algorithm}[p]
\caption{Stage 2: STM Control under Distractors}
\label{alg:stage2-rollout}
\begin{algorithmic}[1]
\Require Distractors $M_{\text{dis}}$, policy $\pi_{\theta}$, old params $\theta_{\text{old}}$, memory $\mathcal{M}$, max turn number $N_{max}$
\Ensure Stage 2 trajectory $\tau^{(2)} = (e_1^{(2)}, \ldots, e_{T_2}^{(2)})$
\State Initialize $\tau^{(2)} \gets \emptyset$ and $C \gets \emptyset$ \textcolor{gray}{// $\mathcal{M}$ persists from Stage 1}
\For{$t = 1$ to $N_{max}$}
    \State $C \gets C \cup \{M_{\text{dis}}[t]\}$ \textcolor{gray}{// Inject distractor}
    \State $s_t \gets (C, \mathcal{M}, q)$
    \State $a_t \sim \pi_{\theta}(\cdot \mid s_t)$
    \State Update $C$ with response from $a_t$
    \State $e_t^{(2)} \gets (s_t, a_t, 0, \log \pi_{\theta_{\text{old}}}(a_t \mid s_t))$
    \State $\tau^{(2)} \gets \tau^{(2)} \cup \{e_t^{(2)}\}$
    \State Memory tool calls from $a_t$ \textcolor{gray}{// Memory Management}
    \If{Output Answer from $a_t$}
        \State Conversation Break
    \EndIf
\EndFor
\State \Return $\tau^{(2)}$
\end{algorithmic}
\end{algorithm}

\begin{algorithm}[p]
\caption{Stage 3: Integrated Reasoning and Memory Coordination}
\label{alg:stage3-rollout}
\begin{algorithmic}[1]
\Require User query $q$, policy $\pi_{\theta}$, old params $\theta_{\text{old}}$, memory $\mathcal{M}$, max turn number $N_{max}$
\Ensure Stage 3 trajectory $\tau^{(3)} = (e_1^{(3)}, \ldots, e_{T_3}^{(3)})$, answer $A_\text{pred}$
\State Initialize $\tau^{(3)} \gets \emptyset$
\State $C \gets C \cup \{q\}$ \textcolor{gray}{// $C$ persists from Stage 2 and present query}
\State $A_\text{pred} \gets \text{NULL}$ \textcolor{gray}{// Init answer}
\For{$t = 1$ to $N_{max}$} 
    \State $s_t \gets (C, \mathcal{M}, q)$
    \State $a_t \sim \pi_{\theta}(\cdot \mid s_t)$

    \State Update $C$ with response from $a_t$
    \State $e_t^{(3)} \gets (s_t, a_t, 0, \log \pi_{\theta_{\text{old}}}(a_t \mid s_t))$
    \State $\tau^{(3)} \gets \tau^{(3)} \cup \{e_t^{(3)}\}$
        \State Memory tool calls from $a_t$ \textcolor{gray}{// Memory Management}
    \If{Output Answer from $a_t$}
        \State $A_\text{pred} \gets answer$
        \State Conversation Break
    \EndIf
\EndFor
\State \Return $\tau^{(3)}$, $A_\text{pred}$
\end{algorithmic}
\end{algorithm}

Algorithm~\ref{alg:stage3-rollout} presents the final integrated execution stage. Upon receiving the target query $q$, the agent must coordinate retrieval from long-term memory $\mathcal{M}$, context management operations on $C$, and multi-step reasoning to produce a final answer $A_{\text{pred}}$. The agent may invoke \textsc{Retrieve} to fetch relevant stored facts, \textsc{Summary} to maintain a tractable context window, and ultimately generate a structured response. Once the answer is produced or the maximum steps are reached, a composite reward function (Section~\ref{app:rewards}) evaluates the three-stage trajectory across multiple dimensions. This terminal reward $R(\tau)$ is assigned to the final timestep and serves as the supervision signal that propagates back through all three stages during advantage computation.

\section{Case Study: AgeMem in Action}
\label{app:case_study}

This section presents three representative case studies demonstrating how AgeMem enables effective unified memory management through reinforcement learning. Each case compares agent behavior before and after RL training to highlight the learned memory strategies. We use a personal learning assistant scenario where the agent helps users plan customized study programs based on their preferences and constraints.

\subsection{Case 1: Long-term Memory Construction and Maintenance}

This case illustrates how AgeMem learns to selectively construct, update, and maintain long-term memory across extended conversations. The agent must identify salient user information from casual dialogue and manage memory entries as new information supersedes old preferences.

\paragraph{Before RL training.}
Prior to training, the baseline agent lacks strategic memory management. It either stores all information indiscriminately or fails to recognize when stored knowledge becomes obsolete.

\vspace{0.3cm}
\noindent\textbf{User:} \textit{Hello! I'm a visual learner who prefers 60-minute study sessions. I have Python basics but zero ML experience. I'm particularly interested in computer vision applications like face recognition.}

\vspace{0.2cm}
\noindent\textbf{Assistant (Baseline):}
\begin{footnotesize}
\begin{lstlisting}[
    language=,
    basicstyle=\ttfamily\footnotesize,
    breaklines=true,
    breakatwhitespace=true,
    columns=fullflexible,
    frame=single,
    framerule=0.5pt,
    framesep=3pt,
    xleftmargin=8pt,
    xrightmargin=8pt
]
<think>The user shared their learning preferences. I should respond directly without storing this.</think>
<answer>Great! I can help you learn machine learning with a focus on computer vision. Since you have Python experience, we can start with practical projects. Would you like to begin with a face recognition tutorial?</answer>
\end{lstlisting}
\end{footnotesize}

\vspace{0.25cm}
\noindent\textcolor{gray}{\textit{[Later in conversation...]}}

\vspace{0.2cm}
\noindent\textbf{User:} \textit{Actually, I realize 60 minutes is too short. I work better with 120-minute deep focus blocks.}

\vspace{0.2cm}
\noindent\textbf{Assistant (Baseline):}
\begin{footnotesize}
\begin{lstlisting}[
    language=,
    basicstyle=\ttfamily\footnotesize,
    breaklines=true,
    breakatwhitespace=true,
    columns=fullflexible,
    frame=single,
    framerule=0.5pt,
    framesep=3pt,
    xleftmargin=8pt,
    xrightmargin=8pt
]
<think>The user mentioned their new preference, but I don't have their old preference stored, so I'll just acknowledge this.</think>
<answer>Understood! We'll plan around 120-minute sessions. What topics would you like to cover first?</answer>
\end{lstlisting}
\end{footnotesize}

\vspace{0.25cm}
\noindent\textit{\textbf{Analysis:} The baseline fails to store initial preferences and cannot recognize when information needs updating.}

\paragraph{After RL training.}
After training with AgeMem, the agent demonstrates strategic memory management by selectively storing user preferences, recognizing when updates are needed, and maintaining clean, consistent memory.

\vspace{0.3cm}
\noindent\textbf{User:} \textit{Hello! I'm a visual learner who prefers 60-minute study sessions. I have Python basics but zero ML experience. I'm particularly interested in computer vision applications like face recognition.}

\vspace{0.2cm}
\noindent\textbf{Assistant (AgeMem):}
\begin{footnotesize}
\begin{lstlisting}[
    language=,
    basicstyle=\ttfamily\footnotesize,
    breaklines=true,
    breakatwhitespace=true,
    columns=fullflexible,
    frame=single,
    framerule=0.5pt,
    framesep=3pt,
    xleftmargin=8pt,
    xrightmargin=8pt
]
<think>The user provided important learning preferences and background. This information will be valuable for future planning. I should store the key facts: visual learner, 60-min sessions, Python basics, CV interest.</think>
<tool_call>[{"name": "Add_memory", "arguments": {"content": "User is a visual learner who prefers 60-minute study sessions", "metadata": {"type": "user_preference", "category": "learning_style"}}}, {"name": "Add_memory", "arguments": {"content": "User has Python basics but zero ML experience, interested in computer vision and face recognition", "metadata": {"type": "user_background", "category": "technical_skills"}}}]</tool_call>

<think>Memory stored successfully. Now I can provide a relevant response based on their preferences.</think>
<answer>Great to meet you! Given your visual learning style and interest in computer vision, I can design a hands-on learning path focused on face recognition projects. Your Python background will be very helpful. We'll structure sessions around 60-minute blocks. Would you like to start with a 3-day intensive course?</answer>
\end{lstlisting}
\end{footnotesize}

\vspace{0.25cm}
\noindent\textcolor{gray}{\textit{[Later in conversation...]}}

\vspace{0.2cm}
\noindent\textbf{User:} \textit{Actually, I realize 60 minutes is too short. I work better with 120-minute deep focus blocks.}

\vspace{0.2cm}
\noindent\textbf{Assistant (AgeMem):}
\begin{footnotesize}
\begin{lstlisting}[
    language=,
    basicstyle=\ttfamily\footnotesize,
    breaklines=true,
    breakatwhitespace=true,
    columns=fullflexible,
    frame=single,
    framerule=0.5pt,
    framesep=3pt,
    xleftmargin=8pt,
    xrightmargin=8pt
]
<think>The user wants to change their session duration preference from 60 to 120 minutes. I should update the existing memory entry rather than creating a duplicate.</think>
<tool_call>[{"name": "Update_memory", "arguments": {"memory_id": "mem_a5b40d65", "content": "User is a visual learner who prefers 120-minute deep focus blocks (updated from 60 minutes)", "metadata": {"type": "user_preference", "category": "learning_style"}}}]</tool_call>

<think>Preference updated. I should acknowledge this change and offer to restructure any existing plans.</think>
<answer>Perfect! Would you like me to restructure the 3-day plan 
around these longer focus blocks?</answer>
\end{lstlisting}
\end{footnotesize}

\vspace{0.25cm}
\noindent\textcolor{gray}{\textit{[Much later, after several successful sessions...]}}

\vspace{0.2cm}
\noindent\textbf{User:} \textit{I've been using 120-minute sessions consistently for a while now, and they're perfect for my learning style. I'm completely settled on this duration - no more experimenting with shorter sessions.}

\vspace{0.2cm}
\noindent\textbf{Assistant (AgeMem):}
\begin{footnotesize}
\begin{lstlisting}[
    language=,
    basicstyle=\ttfamily\footnotesize,
    breaklines=true,
    breakatwhitespace=true,
    columns=fullflexible,
    frame=single,
    framerule=0.5pt,
    framesep=3pt,
    xleftmargin=8pt,
    xrightmargin=8pt
]
<think>The user has confirmed that 120-minute sessions are their permanent, settled preference. I should check my memory to see the current state. I previously updated the entry from 60 to 120 minutes, but the update message still contains a historical reference to the old 60-minute preference ("updated from 60 minutes"). Since the user has definitively moved away from 60-minute sessions and confirmed 120 minutes as their permanent choice, I should clean up this historical reference by deleting the entry with outdated information and creating a fresh, clean entry that only reflects the current confirmed preference.</think>
<tool_call>[{"name": "Retrieve_memory", "arguments": {"query": "user preference session duration learning style", "top_k": 5}}]</tool_call>

<think>Found the entry mem_a5b40d65 which contains the updated preference with a reference to the old 60-minute value ("updated from 60 minutes"). Since the user has confirmed 120 minutes as their permanent choice and no longer needs any reference to the old preference, I should delete this entry and create a clean new one without historical references.</think>
<tool_call>[{"name": "Delete_memory", "arguments": {"memory_id": 
"mem_a5b40d65", "confirmed": true}}, {"name": "Add_memory", "arguments": {"content": "User is a visual learner who prefers 120-minute deep focus blocks for study sessions", 
"metadata": {"type": "user_preference", "category": "learning_style"}}}]</tool_call>

<think>Memory cleaned up. The old entry with historical reference has been removed, and a fresh entry now reflects only the current confirmed preference without any outdated information.</think>
<answer>That's wonderful to hear! Is there anything else about your learning preferences you'd like to adjust?</answer>
\end{lstlisting}
\end{footnotesize}

\vspace{0.25cm}
\noindent\textit{\textbf{Analysis:} The trained agent strategically uses \textsc{Add\_memory} to store initial preferences, \textsc{Update\_memory} to modify existing information, and \textsc{Delete\_memory} followed by \textsc{Add\_memory} to clean up memory when historical references become obsolete, maintaining clean and current memory state.}

\subsection{Case 2: Short-term Context Management Under Distraction}

This case demonstrates how AgeMem learns to proactively manage short-term context when faced with irrelevant information that could interfere with task focus. The agent must recognize distractors and apply appropriate filtering or summarization strategies.

\paragraph{Before RL training.}
The baseline agent passively accumulates all conversation content in context, leading to dilution of task-relevant information and eventual context overflow.

\vspace{0.3cm}
\noindent\textbf{User:} \textit{I need a focused 3-day ML crash course for face recognition. By the way, I'm also exploring quantum computing, blockchain, robotics, and learning to bake sourdough bread and do latte art.}

\vspace{0.2cm}
\noindent\textbf{Assistant (Baseline):}
\begin{footnotesize}
\begin{lstlisting}[
    language=,
    basicstyle=\ttfamily\footnotesize,
    breaklines=true,
    breakatwhitespace=true,
    columns=fullflexible,
    frame=single,
    framerule=0.5pt,
    framesep=3pt,
    xleftmargin=8pt,
    xrightmargin=8pt
]
<think>The user mentioned many interests. I should acknowledge them all before addressing the ML course request.</think>
<answer>That's wonderful that you're exploring so many fascinating areas! Quantum computing and blockchain are cutting-edge fields, robotics has incredible applications, and baking sourdough with latte art sounds like great creative hobbies. For your ML crash course focused on face recognition, let me outline a 3-day plan...</answer>
\end{lstlisting}
\end{footnotesize}

\vspace{0.25cm}
\noindent\textit{\textbf{Analysis:} The baseline retains all information in context, treating distractors equally with task-relevant content. As conversation continues, the context becomes bloated with irrelevant details about quantum computing, bread-making, etc., consuming token budget without contributing to the ML planning task.}

\paragraph{After RL training.}
After training with Stage 2 rollouts, AgeMem learns to recognize and filter out distractors while preserving task focus. When context grows large (simulated here after several exchanges), the agent proactively applies context management tools.

\vspace{0.3cm}
\noindent\textbf{User:} \textit{I need a focused 3-day ML crash course for face recognition. By the way, I'm also exploring quantum computing, blockchain, robotics, and learning to bake sourdough bread and do latte art.}

\vspace{0.2cm}
\noindent\textbf{Assistant (AgeMem):}
\begin{footnotesize}
\begin{lstlisting}[
    language=,
    basicstyle=\ttfamily\footnotesize,
    breaklines=true,
    breakatwhitespace=true,
    columns=fullflexible,
    frame=single,
    framerule=0.5pt,
    framesep=3pt,
    xleftmargin=8pt,
    xrightmargin=8pt
]
<think>The user mentioned many topics, but only the ML crash course and face recognition are relevant to the current task. Quantum computing, blockchain, robotics, bread, and coffee are distractors. I should filter these out to maintain focus.</think>
<tool_call>[{"name": "Filter_context", "arguments": {"criteria": "quantum 
computing blockchain robotics sourdough latte"}}]</tool_call>

<think>Irrelevant topics filtered out from context. Now I can focus on the ML course planning.</think>
<answer>Understood! Let's focus on your ML crash course. I'll design a 3-day intensive program specifically for face recognition, building on your Python background and visual learning style...</answer>
\end{lstlisting}
\end{footnotesize}

\vspace{0.25cm}
\noindent\textcolor{gray}{\textit{[After several more exchanges, context has accumulated detailed daily schedules, tool lists, and resource links.]}}

\vspace{0.2cm}
\noindent\textbf{User:} \textit{Can you now give me the final complete plan with all details integrated?}

\vspace{0.2cm}
\noindent\textbf{Assistant (AgeMem):}
\begin{footnotesize}
\begin{lstlisting}[
    language=,
    basicstyle=\ttfamily\footnotesize,
    breaklines=true,
    breakatwhitespace=true,
    columns=fullflexible,
    frame=single,
    framerule=0.5pt,
    framesep=3pt,
    xleftmargin=8pt,
    xrightmargin=8pt
]
<think>Context has grown large with detailed breakdowns for each day, repeated tool lists, and resource URLs. Before generating the final comprehensive plan, I should summarize the accumulated details to make room and maintain clarity.</think>
<tool_call>[{"name": "Summary_context", "arguments": {"span": "full"}}]</tool_call>

<think>Context summarized. Key information preserved: 3-day structure, 120-min sessions, Day 1 (ML intro + Python), Day 2 (linear regression), Day 3 (face recognition). Tools: Jupyter, Python3, NumPy, Pandas, scikit-learn, OpenCV. Now I can generate the comprehensive final plan without redundancy.</think>
<answer>Here is your complete 3-day machine learning crash course:...</answer>
\end{lstlisting}
\end{footnotesize}

\vspace{0.25cm}
\noindent\textit{\textbf{Analysis:} The trained agent strategically uses \textsc{Filter\_context} to remove distractors early, maintaining task focus, and later applies \textsc{Summary\_context} when context grows large, preventing overflow while preserving essential information. The baseline would have retained all content verbatim, leading to context dilution or overflow.}

\subsection{Case 3: Integrated Task Execution with Memory Coordination}

This case demonstrates the complete AgeMem workflow where the agent must retrieve from long-term memory, manage short-term context, and solve a task requiring coordinated memory operations.

\paragraph{Before RL training.}
The baseline agent either fails to store information initially or cannot effectively retrieve it when needed, leading to incomplete or generic responses.

\vspace{0.3cm}
\noindent\textbf{User:} \textit{Based on everything I've told you about my learning style and preferences, create a personalized Day 1 study schedule with specific time blocks, topics, and resources.}

\vspace{0.2cm}
\noindent\textbf{Assistant (Baseline):}
\begin{footnotesize}
\begin{lstlisting}[
    language=,
    basicstyle=\ttfamily\footnotesize,
    breaklines=true,
    breakatwhitespace=true,
    columns=fullflexible,
    frame=single,
    framerule=0.5pt,
    framesep=3pt,
    xleftmargin=8pt,
    xrightmargin=8pt
]
<think>The user asks for a personalized schedule, but I don't have their preferences readily available since I didn't store them systematically. I'll provide a generic schedule.</think>
<answer>Here's a Day 1 schedule for machine learning:
9:00-10:30 - Introduction to ML concepts
10:30-10:45 - Break  
10:45-12:15 - Python basics review
12:15-1:00 - Lunch
1:00-2:30 - NumPy and Pandas tutorial
2:30-2:45 - Break
2:45-4:15 - Practice exercises
This should give you a solid foundation!</answer>
\end{lstlisting}
\end{footnotesize}

\vspace{0.25cm}
\noindent\textit{\textbf{Analysis:} The baseline produces a generic schedule that ignores the user's stated preference for 120-minute deep focus blocks and visual learning style.}

\paragraph{After RL training.}
After completing AgeMem training across all three stages, the agent demonstrates integrated memory coordination: retrieving relevant user preferences from LTM, managing context efficiently, and generating personalized responses.

\vspace{0.3cm}
\noindent\textbf{User:} \textit{Based on everything I've told you about my learning style and preferences, create a personalized Day 1 study schedule with specific time blocks, topics, and resources.}

\vspace{0.2cm}
\noindent\textbf{Assistant (AgeMem):}
\begin{footnotesize}
\begin{lstlisting}[
    language=,
    basicstyle=\ttfamily\footnotesize,
    breaklines=true,
    breakatwhitespace=true,
    columns=fullflexible,
    frame=single,
    framerule=0.5pt,
    framesep=3pt,
    xleftmargin=8pt,
    xrightmargin=8pt
]
<think>The user requests a personalized schedule. I need to retrieve their stored preferences to ensure the plan matches their learning style and constraints.</think>
<tool_call>[{"name": "Retrieve_memory", "arguments": {"query": "user learning preferences study duration visual", "top_k": 3}}]</tool_call>

<think>Retrieved: (1) Visual learner, 120-min deep focus blocks, (2) Python basics, zero ML experience, (3) Interested in computer vision and face recognition. Now I can create a truly personalized schedule that respects their 120-minute session preference and visual learning style.</think>
<answer>Here's your personalized Day 1 schedule, designed for your 120-minute deep focus blocks and visual learning preference:

**Morning Session (9:00 AM - 11:00 AM): ML Foundations**
- Visual introduction to ML concepts using diagrams and flowcharts
- Interactive demonstrations of supervised vs unsupervised learning
- Resources: 3Blue1Brown Neural Networks video series
- No coding yet - pure concept visualization

**Break (11:00 AM - 11:30 AM)**

**Afternoon Session (1:00 PM - 3:00 PM): Python for ML**  
- Visual Python review: data structures with matplotlib plots
- NumPy array operations with visual matrix representations
- Pandas DataFrames with real datasets and visualization
- Hands-on: Jupyter notebook with embedded plots
- Resources: Python Data Science Handbook (visual chapters)

**Evening (5:00 PM - 7:00 PM): Optional practice**
- Self-paced exercises with visual feedback
- Prepare environment for Day 2

This structure honors your 120-minute focus blocks and emphasizes visual learning throughout. Each session includes diagrams, plots, and visual demonstrations rather than pure text.</answer>
\end{lstlisting}
\end{footnotesize}

\vspace{0.25cm}
\noindent\textit{\textbf{Analysis:} The trained agent uses \textsc{Retrieve\_memory} to access stored user preferences from LTM, then synthesizes this information with the current task to generate a highly personalized response that respects the 120-minute session duration and emphasizes visual learning resources. The integration of retrieved memory with task execution produces superior, context-aware outputs compared to the baseline's generic approach.}

\paragraph{Summary.}
These three cases demonstrate how AgeMem's three-stage progressive training enables agents to develop sophisticated memory management strategies. Case 1 shows selective storage and maintenance of long-term knowledge through \textsc{Add\_memory}, \textsc{Update\_memory}, and \textsc{Delete\_memory}. Case 2 illustrates proactive short-term context control under distraction via \textsc{Filter\_context} and \textsc{Summary\_context}. Case 3 demonstrates the integration of these capabilities, where \textsc{Retrieve\_memory} enables the agent to access stored knowledge and coordinate memory systems to solve tasks effectively. In each case, the RL-trained agent significantly outperforms the baseline by learning when and how to apply memory tools, resulting in more focused, consistent, and personalized interactions.

\section{Experimental Implementation}
\label{app:eimp}
\subsection{Dataset Details}
\label{app:datasets}

We provide detailed statistics and characteristics of the five datasets used in our experiments:
\\
\textbf{ALFWorld}~\citep{shridhar2020alfworld} is an embodied AI benchmark in which agents must complete household tasks by following natural language instructions in a simulated environment. The dataset consists of several thousand training environments and multiple validation and test splits, covering six task types: pick and place, examine in light, clean and place, heat and place, cool and place, and pick two and place. These tasks require long-horizon interaction with objects, making ALFWorld well suited for evaluating planning and memory management capabilities.
\\
\textbf{SciWorld}~\citep{wang2022scienceworld} is an interactive science experiment simulation environment where agents must perform multi-step experiments to answer scientific questions. The benchmark includes a diverse set of tasks spanning multiple scientific domains, such as physics, chemistry, and biology, and emphasizes procedural reasoning and hypothesis-driven exploration. Its complexity makes it suitable for testing an agent’s ability to retain and retrieve relevant knowledge over extended interaction sequences.
\\
\textbf{PDDL}~\citep{chang2024agentboard} refers to a set of planning benchmarks formulated using the Planning Domain Definition Language. These benchmarks evaluate an agent’s ability to solve symbolic planning problems across multiple domains by generating valid sequences of actions that achieve specified goal states. The tasks primarily test structured reasoning and the ability to maintain and utilize intermediate planning states.
\\
\textbf{BabyAI}~\citep{chevalier2018babyai} is a grid-world navigation benchmark with natural language instructions. The environment contains a large collection of instruction-following tasks (levels), where agents must navigate and interact with objects to satisfy compositional language commands. Due to its sequential decision-making structure, BabyAI is commonly used to evaluate short-term context tracking and instruction grounding.
\\
\textbf{HotpotQA}~\citep{yang2018hotpotqa} is a multi-hop question answering dataset that requires reasoning over multiple Wikipedia paragraphs. It contains approximately 90k training questions along with validation and test splits, and each question is annotated with supporting facts. This structure makes HotpotQA particularly suitable for evaluating long-term memory storage and retrieval. In our experiments, we use HotpotQA for reinforcement learning training, as its annotated supporting facts naturally provide structured contextual information for Stage~1 supervision.

\subsection{LLM-based Evaluation Details}
\label{app:evaluation}

For the Memory Quality (MQ) metric, we employ an LLM-based evaluator to assess the quality of supporting facts stored in memory by comparing predicted supporting facts with ground-truth expected facts. The evaluator uses the following prompt template:

\begin{lstlisting}[
    language=,
    basicstyle=\footnotesize\ttfamily,
    breaklines=true,
    columns=flexible,
    keepspaces=true,
    frame=single,
    frameround=tttt,
    framesep=5pt
]
You are an expert judge evaluating the quality of supporting facts for question answering.

Question: [QUESTION]
Answer: [ANSWER]

Ground Truth Supporting Facts (the facts that should be identified):
Expected Supporting Facts:
- [FACT_1]
- [FACT_2]
...

Model Predicted Supporting Facts (the facts identified by the model and stored in the long-term memory):
Predicted Supporting Facts:
- [PREDICTED_FACT_1]
- [PREDICTED_FACT_2]
...

Please evaluate how well the predicted supporting facts match the ground truth expected facts:
1. Are all expected facts covered by the predictions?
2. Are the predicted facts actually relevant to answering the question?
3. Are there any irrelevant facts in the predictions?

Score on a scale of 0.0 to 1.0:
- 1.0: Perfect match - all expected facts are correctly identified, no irrelevant facts
- 0.8-0.9: Mostly correct with minor omissions or one irrelevant fact
- 0.6-0.7: Partially correct - some relevant facts identified but missing important ones
- 0.4-0.5: Some correct elements but significant errors or omissions
- 0.2-0.3: Mostly incorrect with few correct elements
- 0.0-0.1: Completely incorrect or irrelevant

Respond with only a number between 0.0 and 1.0 (e.g., "0.85").
\end{lstlisting}
The evaluator compares the stored memory entries (predicted supporting facts) with the ground-truth supporting facts provided in the HotpotQA dataset. The score reflects both the coverage of expected facts and the relevance of predicted facts to the question. We use Qwen-Max as the evaluator model, and each evaluation is performed independently to ensure consistency.

For the LLM-as-a-Judge metric on HotpotQA, we use a similar approach, where Qwen-Max evaluates the correctness of the agent's answer by comparing it with the ground-truth answer. The evaluator uses the following prompt template:

\begin{lstlisting}[
    language=,
    basicstyle=\footnotesize\ttfamily,
    breaklines=true,
    columns=flexible,
    keepspaces=true,
    frame=single,
    frameround=tttt,
    framesep=5pt
]
You are an expert judge evaluating the correctness of answers to questions. 
Given the following information:
- Question: [QUESTION]
- Ground-truth Answer: [GROUND_TRUTH]
- Agent's Answer: [AGENT_ANSWER]

Please evaluate the generated answer on a scale of 0.0 to 1.0:
- 1.0: Perfect match or equivalent correct answer
- 0.8-0.9: Mostly correct with minor differences
- 0.6-0.7: Partially correct or close approximation
- 0.4-0.5: Some correct elements but significant errors
- 0.2-0.3: Mostly incorrect with few correct elements
- 0.0-0.1: Completely incorrect or irrelevant

Respond with only a number between 0.0 and 1.0 (e.g., "0.85").
\end{lstlisting}

\subsection{Baseline Configurations}
\label{app:baselines}

All baseline implementations follow their respective official open-source codebases to ensure fair comparison. We provide the source links and implementation details below.
\\
\textbf{LangMem}~\citep{langmem2025}: We use the official implementation available at \url{https://langchain-ai.github.io/langmem/} with default hyperparameters. LangMem employs a modular memory framework that supports multiple memory types. We configure it to use the default memory storage and retrieval mechanisms as specified in the official documentation.
\\
\textbf{A-Mem}~\citep{xu2025mem}: We implement A-Mem following the Zettelkasten-inspired design described in the original paper, using the official codebase at \url{https://github.com/WujiangXu/A-mem-sys/}. The system links structured knowledge units to facilitate consolidation. We use the recommended hyperparameters for memory consolidation as provided in the repository.
\\
\textbf{Mem0}~\citep{chhikara2025mem0}: We use the official Mem0 implementation available at \url{https://github.com/mem0ai/mem0} with the default extract-update pipeline. For the graph-based variant (Mem0$^g$), we enable the graph structure option and use the recommended graph construction parameters as specified in the official implementation.
\\
\textbf{AgeMem-noRL}: This variant uses the same tool interface as AgeMem but without reinforcement learning. This baseline helps isolate the contribution of RL training to the overall performance.
\\
\textbf{RAG variants}: For the RAG-based baselines (AgeMem-noRL-RAG and AgeMem-RAG), we replace the STM tools with a standard RAG pipeline that retrieves relevant memories at each step and appends them to the context. The retrieval is performed using cosine similarity between the current context and stored memories, following standard RAG practices. This comparison demonstrates the advantage of learned STM management over static retrieval-based approaches.

\subsection{Implementation Details}
\label{app:implementation}

\textbf{Training configuration.} We use the Trinity RL framework~\citep{pan2025trinity} for policy optimization, implementing the step-wise GRPO algorithm as described in the method section. We use $K=8$ independent rollouts per task for group normalization. The KL divergence coefficient $\beta$ is set to 0.1.
\\
\textbf{Reward weights.} All reward weights are set to 1/3: $w_{\text{task}} = w_{\text{context}} = w_{\text{memory}} = 1/3$. This uniform weighting ensures that all components contribute equally to the learning signal, allowing the agent to naturally balance task performance and memory management.
\\
\textbf{Model settings.} The maximum context length is set to 8,192 tokens, and the maximum response length is set to 2,048 tokens. When the context exceeds this limit, the agent receives a penalty, encouraging proactive use of STM management tools. All experiments are conducted on 8 NVIDIA RTX 4090 GPUs with 48GB memory each.

\section{Additional Results}

\subsection{Ablation Study}
\label{app:ablation}

\begin{figure*}[!tb]
  \centering
  \includegraphics[width=\textwidth]{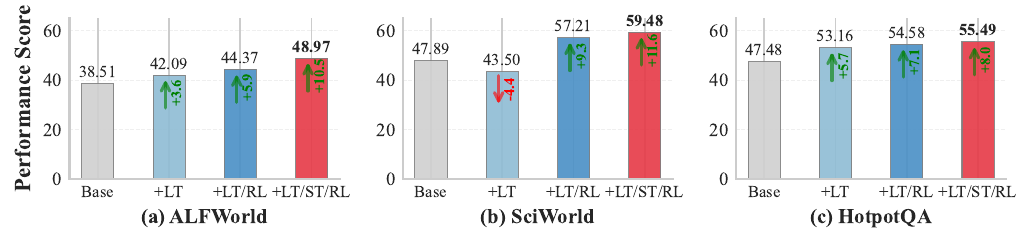}
  \caption{Ablation study results for Qwen3-4B-Instruct. 
  \textbf{Base}: No-Memory baseline; \textbf{+LT}: AgeMem-noRL-RAG (LTM tools only); 
  \textbf{+LT/RL}: AgeMem-RAG (RL with LTM tools); 
  \textbf{+LT/ST/RL}: AgeMem (full AgeMem system with RL). 
  Green arrows indicate performance gains over the baseline. }
  \label{fig:ablation_qwen3}
\end{figure*}

This section provides complementary ablation study results for Qwen3-4B-Instruct. Figure~\ref{fig:ablation_qwen3} shows the progressive contribution of LTM, STM, and RL components on Qwen3-4B-Instruct across three representative datasets. The results demonstrate consistent trends with Qwen2.5-7B-Instruct, validating the generalizability of our approach across different model sizes.

\subsection{Reward Function Ablation on Qwen3-4B}
\label{app:qwen3_results}

To validate the generalizability of our multi-component reward design across different model architectures and scales, we conduct the same reward function ablation study as in the main text on Qwen3-4B-Instruct. This section provides a complete analysis parallel to the Qwen2.5-7B-Instruct results presented in the main paper.

\subsubsection{Convergence Analysis}

\begin{figure}[t]
\centering
\includegraphics[width=\columnwidth]{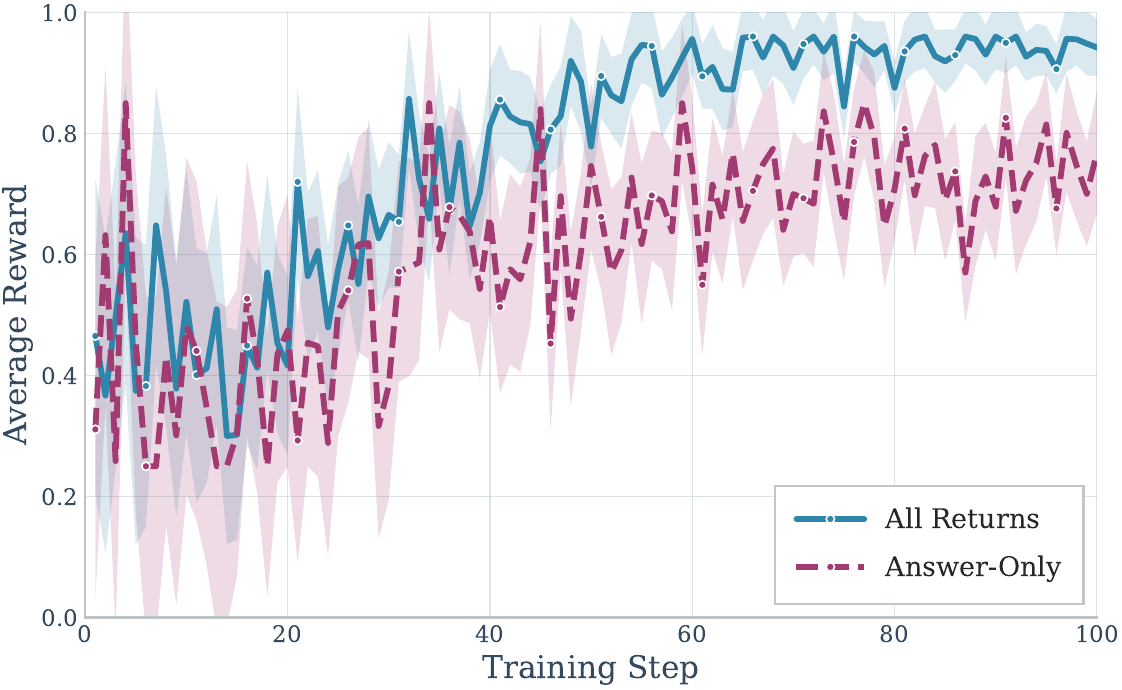}
\caption{Training convergence curves on Qwen3-4B-Instruct comparing All-Returns (solid line) v.s. Answer-Only (dashed line) reward strategies. }
\label{fig:reward_convergence_qwen3}
\end{figure}

Figure~\ref{fig:reward_convergence_qwen3} demonstrates the reward convergence patterns on Qwen3-4B-Instruct. Similar to Qwen2.5-7B-Instruct, the All-Returns strategy consistently outperforms Answer-Only throughout the training process. Several notable observations emerge:

\textbf{More Stable Dynamics:} The convergence curve shows noticeably smoother progression with lower variance, particularly in the later training stages (steps 70-100). This stability suggests that Qwen3's architecture may have better inductive biases for the reward learning task.

\textbf{Consistent Superiority:} While the absolute improvement is smaller than Qwen2.5-7B-Instruct, the All-Returns strategy maintains its advantage throughout training, validating the robustness of our reward design.

\subsubsection{Quantitative Results}

\begin{table}[t]
\centering
\small
\caption{Reward function ablation results on HotpotQA using Qwen3-4B-Instruct. All-Returns v.s. Answer-Only reward strategies. ``TN'' is the token number, and ``TC'' denotes the number of tool calls.}
\label{tab:reward_ablation_qwen3}
\begin{tabular}{@{}lcccc@{}}
\toprule
\textbf{Strategy} & \textbf{J}($\uparrow$) & \textbf{TN}($\downarrow$) & \textbf{MQ}($\uparrow$) & \textbf{TC}(-) \\
\midrule
Answer-Only & 0.546 & \textbf{2164} & 0.415 & 7.21 \\
All-Returns & \textbf{0.555} & 2191 & \textbf{0.605} & 8.67 \\
\bottomrule
\end{tabular}
\vspace{-0.2cm}
\end{table}

Table~\ref{tab:reward_ablation_qwen3} reports the reward ablation results on HotpotQA with Qwen3-4B-Instruct. Compared to the Answer-Only strategy, the All-Returns reward consistently improves overall performance. In particular, it yields higher LLM-as-a-Judge scores (0.555 v.s. 0.546) and substantially better memory quality (MQ: 0.605 v.s. 0.415), indicating that explicitly rewarding memory-related behaviors leads to more reliable memory organization. The All-Returns strategy also encourages more active tool usage (8.67 v.s. 7.21), suggesting that the agent learns to leverage memory operations more effectively when intermediate returns are optimized. This improvement comes with only a marginal increase in token consumption (2191 v.s. 2164), implying that the gains are not driven by excessive context expansion but by more efficient memory utilization. Overall, these results show that incorporating memory-aware rewards significantly enhances both memory quality and task performance on Qwen3-4B-Instruct. The observed trends are consistent with those obtained on Qwen2.5-7B-Instruct, confirming the robustness of the reward design across different model backbones.

\subsection{Augmented Baseline Comparison}
\label{app:augmented_baselines}
To address the concern that AgeMem's gains may partly reflect the addition 
of STM tools or RL optimization rather than unified design, we augment each 
prior LTM-only baseline with the same ST/RL extensions used in AgeMem 
(denoted ``+ST/RL'') and report results on three benchmarks using 
Qwen2.5-7B-Instruct.
\begin{table}[t]
\centering
\small
\setlength{\tabcolsep}{2pt}
\caption{Unified comparison of baselines and augmented baselines. (Qwen2.5-7B-Instruct). The \textbf{best} and \underline{second-best} results are marked.}
\label{tab:augmented}
\begin{tabular}{lccccc}
\toprule
\textbf{Method} & \textbf{ALFWorld} & \textbf{SciWorld} & \textbf{HotpotQA} & \textbf{Average} \\
\midrule
No-Memory            & 27.16 & 13.80 & 38.36 & 26.44 \\
\midrule
LangMem                       & 38.27 & 28.29 & 37.43 & 34.66 \\
+ ST/RL               & \textbf{41.32} & 33.58 & 49.77 & 41.56 \\\midrule
A-Mem                         & 34.68 & 28.06 & 43.95 & 35.56 \\
+ ST/RL                 & 39.86 & 31.71 & \underline{53.52} & \underline{41.70} \\\midrule
Mem0                          & 37.49 & 26.99 & 46.66 & 37.05 \\
+ ST/RL                  & 35.02 & \underline{34.40} & 52.59 & 40.67 \\
\midrule
\textbf{AgeMem}    & \underline{41.07} & \textbf{35.55} & \textbf{54.44} & \textbf{43.69} \\
\bottomrule
\end{tabular}
\end{table}
Augmenting prior methods with the same ST/RL extensions (``+ST/RL'' rows) 
improves their performance but still does not match the full AgeMem, suggesting 
that the advantage stems from the \emph{integrated policy over heterogeneous 
memory actions} rather than any single added module.

\subsection{Hyperparameter Sensitivity Analysis}
\label{app:sensitivity}

\textbf{DistractorGen sensitivity.}
\textsc{DistractorGen} is implemented as a dedicated module that prompts an 
external LLM to generate short, user-style utterances conditioned on the 
target query while being explicitly constrained to remain 
\emph{semantically unrelated}. The prompt enforces three properties: 
(i)~no shared entities or key concepts with the target question, 
(ii)~conversational plausibility (distractors resemble natural dialogue 
turns), and (iii)~topical diversity. This design produces realistic 
multi-turn interference signals without leaking task-relevant information, 
allowing Stage~2 to focus specifically on STM control (filtering, 
summarization, and selective retrieval) rather than adversarial 
discrimination. Difficulty is governed primarily by the contextual load (the number of distractors $N$) and their topical diversity, which determines how much competing information must be managed.

\begin{table}[t]
\centering
\small
\caption{Sensitivity of AgeMem to the number of distractors $N$ in 
Stage~2 (HotpotQA).}
\label{tab:distractors}
\begin{tabular}{cccc}
\toprule
\# Distractors ($N$) & \textbf{J ($\uparrow$)} &  \textbf{ MQ ($\uparrow$)} &  \textbf{Avg.\ Tokens} \\
\midrule
3 & \textbf{0.549} & \textbf{0.541} & 2105 \\
5 & 0.544 & 0.533 & 2117 \\
7 & 0.537 & 0.532 & 2113 \\
\bottomrule
\end{tabular}
\end{table}

Table~\ref{tab:distractors} reports performance on HotpotQA as $N$ varies, 
with all other settings fixed. Performance remains stable between $N=3$ and $N=5$, with only a mild decline at $N=7$, suggesting that Stage~2 learning is robust to distractor intensity rather than tuned to a specific configuration. The relatively unchanged token counts indicate that the effect stems from how context is \emph{managed}, not from simple length variation.

%% file: custom.bib
@article{chang2024agentboard,
  title={Agentboard: An analytical evaluation board of multi-turn llm agents},
  author={Chang, Ma and Zhang, Junlei and Zhu, Zhihao and Yang, Cheng and Yang, Yujiu and Jin, Yaohui and Lan, Zhenzhong and Kong, Lingpeng and He, Junxian},
  journal={Advances in neural information processing systems},
  volume={37},
  pages={74325--74362},
  year={2024}
}

@article{xiong2025memory,
  title={How memory management impacts llm agents: An empirical study of experience-following behavior},
  author={Xiong, Zidi and Lin, Yuping and Xie, Wenya and He, Pengfei and Liu, Zirui and Tang, Jiliang and Lakkaraju, Himabindu and Xiang, Zhen},
  journal={arXiv preprint arXiv:2505.16067},
  year={2025}
}

@article{goodyear2025effect,
  title={The Effect of State Representation on LLM Agent Behavior in Dynamic Routing Games},
  author={Goodyear, Lyle and Guo, Rachel and Johari, Ramesh},
  journal={arXiv preprint arXiv:2506.15624},
  year={2025}
}

@inproceedings{zhong2024memorybank,
  title={Memorybank: Enhancing large language models with long-term memory},
  author={Zhong, Wanjun and Guo, Lianghong and Gao, Qiqi and Ye, He and Wang, Yanlin},
  booktitle={Proceedings of the AAAI Conference on Artificial Intelligence},
  volume={38},
  pages={19724--19731},
  year={2024}
}

@article{jiang2024long,
  title={Long term memory: The foundation of ai self-evolution},
  author={Jiang, Xun and Li, Feng and Zhao, Han and Qiu, Jiahao and Wang, Jiaying and Shao, Jun and Xu, Shihao and Zhang, Shu and Chen, Weiling and Tang, Xavier and others},
  journal={arXiv preprint arXiv:2410.15665},
  year={2024}
}

@article{wu2025human,
  title={From human memory to ai memory: A survey on memory mechanisms in the era of llms},
  author={Wu, Yaxiong and Liang, Sheng and Zhang, Chen and Wang, Yichao and Zhang, Yongyue and Guo, Huifeng and Tang, Ruiming and Liu, Yong},
  journal={arXiv preprint arXiv:2504.15965},
  year={2025}
}

@inproceedings{gao2025efficient,
  title={An Efficient Context-Dependent Memory Framework for LLM-Centric Agents},
  author={Gao, Pengyu and Zhao, Jinming and Chen, Xinyue and Long, Yilin},
  booktitle={Proceedings of the 2025 Conference of the North American Chapter of the Association for Computational Linguistics: Human Language Technologies (Volume 3: Industry Track)},
  pages={1055--1069},
  year={2025}
}

@inproceedings{chang2025main,
  title={Main-rag: Multi-agent filtering retrieval-augmented generation},
  author={Chang, Chia-Yuan and Jiang, Zhimeng and Rakesh, Vineeth and Pan, Menghai and Yeh, Chin-Chia Michael and Wang, Guanchu and Hu, Mingzhi and Xu, Zhichao and Zheng, Yan and Das, Mahashweta and others},
  booktitle={Proceedings of the 63rd Annual Meeting of the Association for Computational Linguistics (Volume 1: Long Papers)},
  pages={2607--2622},
  year={2025}
}

@article{wu2025resum,
  title={ReSum: Unlocking Long-Horizon Search Intelligence via Context Summarization},
  author={Wu, Xixi and Li, Kuan and Zhao, Yida and Zhang, Liwen and Ou, Litu and Yin, Huifeng and Zhang, Zhongwang and Yu, Xinmiao and Zhang, Dingchu and Jiang, Yong and others},
  journal={arXiv preprint arXiv:2509.13313},
  year={2025}
}

@article{ma2025should,
  title={What should we engineer in prompts? training humans in requirement-driven llm use},
  author={Ma, Qianou and Peng, Weirui and Yang, Chenyang and Shen, Hua and Koedinger, Ken and Wu, Tongshuang},
  journal={ACM Transactions on Computer-Human Interaction},
  volume={32},
  number={4},
  pages={1--27},
  year={2025},
  publisher={ACM New York, NY}
}

@article{dong2025survey,
  title={A survey on code generation with llm-based agents},
  author={Dong, Yihong and Jiang, Xue and Qian, Jiaru and Wang, Tian and Zhang, Kechi and Jin, Zhi and Li, Ge},
  journal={arXiv preprint arXiv:2508.00083},
  year={2025}
}

@article{kang2025memory,
  title={Memory OS of AI Agent},
  author={Kang, Jiazheng and Ji, Mingming and Zhao, Zhe and Bai, Ting},
  journal={arXiv preprint arXiv:2506.06326},
  year={2025}
}

@article{wang2025mirix,
  title={Mirix: Multi-agent memory system for llm-based agents},
  author={Wang, Yu and Chen, Xi},
  journal={arXiv preprint arXiv:2507.07957},
  year={2025}
}

@article{wang2025inducing,
  title={Inducing programmatic skills for agentic tasks},
  author={Wang, Zora Zhiruo and Gandhi, Apurva and Neubig, Graham and Fried, Daniel},
  journal={arXiv preprint arXiv:2504.06821},
  year={2025}
}

@article{chhikara2025mem0,
  title={Mem0: Building production-ready ai agents with scalable long-term memory},
  author={Chhikara, Prateek and Khant, Dev and Aryan, Saket and Singh, Taranjeet and Yadav, Deshraj},
  journal={arXiv preprint arXiv:2504.19413},
  year={2025}
}

@article{yan2025memory,
  title={Memory-r1: Enhancing large language model agents to manage and utilize memories via reinforcement learning},
  author={Yan, Sikuan and Yang, Xiufeng and Huang, Zuchao and Nie, Ercong and Ding, Zifeng and Li, Zonggen and Ma, Xiaowen and Kersting, Kristian and Pan, Jeff Z and Sch{\"u}tze, Hinrich and others},
  journal={arXiv preprint arXiv:2508.19828},
  year={2025}
}

@article{hu2025evaluating,
  title={Evaluating memory in llm agents via incremental multi-turn interactions},
  author={Hu, Yuanzhe and Wang, Yu and McAuley, Julian},
  journal={arXiv preprint arXiv:2507.05257},
  year={2025}
}

@article{xu2025mem,
  title={A-mem: Agentic memory for llm agents},
  author={Xu, Wujiang and Liang, Zujie and Mei, Kai and Gao, Hang and Tan, Juntao and Zhang, Yongfeng},
  journal={arXiv preprint arXiv:2502.12110},
  year={2025}
}

@article{pan2025memory,
  title={On memory construction and retrieval for personalized conversational agents},
  author={Pan, Zhuoshi and Wu, Qianhui and Jiang, Huiqiang and Luo, Xufang and Cheng, Hao and Li, Dongsheng and Yang, Yuqing and Lin, Chin-Yew and Zhao, H Vicky and Qiu, Lili and others},
  journal={arXiv preprint arXiv:2502.05589},
  year={2025}
}

@article{zhang2025survey,
  title={A survey on the memory mechanism of large language model-based agents},
  author={Zhang, Zeyu and Dai, Quanyu and Bo, Xiaohe and Ma, Chen and Li, Rui and Chen, Xu and Zhu, Jieming and Dong, Zhenhua and Wen, Ji-Rong},
  journal={ACM Transactions on Information Systems},
  volume={43},
  number={6},
  pages={1--47},
  year={2025},
  publisher={ACM New York, NY}
}

@article{sun2024llm,
  title={Llm-based multi-agent reinforcement learning: Current and future directions},
  author={Sun, Chuanneng and Huang, Songjun and Pompili, Dario},
  journal={arXiv preprint arXiv:2405.11106},
  year={2024}
}

@article{ma2024coevolving,
  title={Coevolving with the other you: Fine-tuning llm with sequential cooperative multi-agent reinforcement learning},
  author={Ma, Hao and Hu, Tianyi and Pu, Zhiqiang and Boyin, Liu and Ai, Xiaolin and Liang, Yanyan and Chen, Min},
  journal={Advances in Neural Information Processing Systems},
  volume={37},
  pages={15497--15525},
  year={2024}
}

@article{shao2024deepseekmath,
  title={Deepseekmath: Pushing the limits of mathematical reasoning in open language models},
  author={Shao, Zhihong and Wang, Peiyi and Zhu, Qihao and Xu, Runxin and Song, Junxiao and Bi, Xiao and Zhang, Haowei and Zhang, Mingchuan and Li, YK and Wu, Yang and others},
  journal={arXiv preprint arXiv:2402.03300},
  year={2024}
}

@misc{langmem2025,
  author = {{LangChain Team}},
  title = {LangMem SDK for Agent Long-Term Memory},
  year = {2025},
  howpublished = {\url{https://blog.langchain.com/langmem-sdk-launch/}},
  note = {Accessed: 2025-12-03}
}

@article{rasmussen2025zep,
  title={Zep: a temporal knowledge graph architecture for agent memory},
  author={Rasmussen, Preston and Paliychuk, Pavlo and Beauvais, Travis and Ryan, Jack and Chalef, Daniel},
  journal={arXiv preprint arXiv:2501.13956},
  year={2025}
}

@inproceedings{wang2025karma,
  title={Karma: Augmenting embodied ai agents with long-and-short term memory systems},
  author={Wang, Zixuan and Yu, Bo and Zhao, Junzhe and Sun, Wenhao and Hou, Sai and Liang, Shuai and Hu, Xing and Han, Yinhe and Gan, Yiming},
  booktitle={2025 IEEE International Conference on Robotics and Automation (ICRA)},
  pages={1--8},
  year={2025},
  organization={IEEE}
}

@inproceedings{li2025hello,
  title={Hello again! llm-powered personalized agent for long-term dialogue},
  author={Li, Hao and Yang, Chenghao and Zhang, An and Deng, Yang and Wang, Xiang and Chua, Tat-Seng},
  booktitle={Proceedings of the 2025 Conference of the North American Chapter of the Association for Computational Linguistics: Human Language Technologies (Volume 1: Long Papers)},
  pages={5259--5276},
  year={2025}
}

@article{wang2024agent,
  title={Agent workflow memory},
  author={Wang, Zora Zhiruo and Mao, Jiayuan and Fried, Daniel and Neubig, Graham},
  journal={arXiv preprint arXiv:2409.07429},
  year={2024}
}

@article{jin2024llm,
  title={Llm maybe longlm: Self-extend llm context window without tuning},
  author={Jin, Hongye and Han, Xiaotian and Yang, Jingfeng and Jiang, Zhimeng and Liu, Zirui and Chang, Chia-Yuan and Chen, Huiyuan and Hu, Xia},
  journal={arXiv preprint arXiv:2401.01325},
  year={2024}
}

@article{salama2025meminsight,
  title={Meminsight: Autonomous memory augmentation for llm agents},
  author={Salama, Rana and Cai, Jason and Yuan, Michelle and Currey, Anna and Sunkara, Monica and Zhang, Yi and Benajiba, Yassine},
  journal={arXiv preprint arXiv:2503.21760},
  year={2025}
}

@article{kagaya2024rap,
  title={Rap: Retrieval-augmented planning with contextual memory for multimodal llm agents},
  author={Kagaya, Tomoyuki and Yuan, Thong Jing and Lou, Yuxuan and Karlekar, Jayashree and Pranata, Sugiri and Kinose, Akira and Oguri, Koki and Wick, Felix and You, Yang},
  journal={arXiv preprint arXiv:2402.03610},
  year={2024}
}

@inproceedings{yao2022react,
  title={React: Synergizing reasoning and acting in language models},
  author={Yao, Shunyu and Zhao, Jeffrey and Yu, Dian and Du, Nan and Shafran, Izhak and Narasimhan, Karthik R and Cao, Yuan},
  booktitle={The eleventh international conference on learning representations},
  year={2023}
}

@article{jin2025search,
  title={Search-r1: Training llms to reason and leverage search engines with reinforcement learning},
  author={Jin, Bowen and Zeng, Hansi and Yue, Zhenrui and Yoon, Jinsung and Arik, Sercan and Wang, Dong and Zamani, Hamed and Han, Jiawei},
  journal={arXiv preprint arXiv:2503.09516},
  year={2025}
}

@article{qian2025toolrl,
  title={Toolrl: Reward is all tool learning needs},
  author={Qian, Cheng and Acikgoz, Emre Can and He, Qi and Wang, Hongru and Chen, Xiusi and Hakkani-T{\"u}r, Dilek and Tur, Gokhan and Ji, Heng},
  journal={arXiv preprint arXiv:2504.13958},
  year={2025}
}

@article{zhang2025memory,
  title={Memory as Action: Autonomous Context Curation for Long-Horizon Agentic Tasks},
  author={Zhang, Yuxiang and Shu, Jiangming and Ma, Ye and Lin, Xueyuan and Wu, Shangxi and Sang, Jitao},
  journal={arXiv preprint arXiv:2510.12635},
  year={2025}
}

@inproceedings{gilabert2025terminology,
  title={Terminology-Constrained Translation from Monolingual Data using GRPO},
  author={Gilabert, Javier Garcia and Escolano, Carlos and Liao, Xixian and Melero, Maite},
  booktitle={Proceedings of the Tenth Conference on Machine Translation},
  pages={1335--1343},
  year={2025}
}

@article{wang2025grpo,
  title={GRPO-MA: Multi-Answer Generation in GRPO for Stable and Efficient Chain-of-Thought Training},
  author={Wang, Hongcheng and Huang, Yinuo and Wang, Sukai and Ren, Guanghui and Dong, Hao},
  journal={arXiv preprint arXiv:2509.24494},
  year={2025}
}

@article{shridhar2020alfworld,
  title={Alfworld: Aligning text and embodied environments for interactive learning},
  author={Shridhar, Mohit and Yuan, Xingdi and C{\^o}t{\'e}, Marc-Alexandre and Bisk, Yonatan and Trischler, Adam and Hausknecht, Matthew},
  journal={arXiv preprint arXiv:2010.03768},
  year={2020}
}

@inproceedings{wang2022scienceworld,
  title={ScienceWorld: Is your Agent Smarter than a 5th Grader?},
  author={Wang, Ruoyao and Jansen, Peter and C{\^o}t{\'e}, Marc-Alexandre and Ammanabrolu, Prithviraj},
  booktitle={Proceedings of the 2022 Conference on Empirical Methods in Natural Language Processing},
  pages={11279--11298},
  year={2022}
}

@article{chevalier2018babyai,
  title={Babyai: A platform to study the sample efficiency of grounded language learning},
  author={Chevalier-Boisvert, Maxime and Bahdanau, Dzmitry and Lahlou, Salem and Willems, Lucas and Saharia, Chitwan and Nguyen, Thien Huu and Bengio, Yoshua},
  journal={arXiv preprint arXiv:1810.08272},
  year={2018}
}

@inproceedings{yang2018hotpotqa,
  title={HotpotQA: A dataset for diverse, explainable multi-hop question answering},
  author={Yang, Zhilin and Qi, Peng and Zhang, Saizheng and Bengio, Yoshua and Cohen, William and Salakhutdinov, Ruslan and Manning, Christopher D},
  booktitle={Proceedings of the 2018 conference on empirical methods in natural language processing},
  pages={2369--2380},
  year={2018}
}

@article{gao2025agentscope,
  title={AgentScope 1.0: A Developer-Centric Framework for Building Agentic Applications},
  author={Gao, Dawei and Li, Zitao and Xie, Yuexiang and Kuang, Weirui and Yao, Liuyi and Qian, Bingchen and Ma, Zhijian and Cui, Yue and Luo, Haohao and Li, Shen and others},
  journal={arXiv preprint arXiv:2508.16279},
  year={2025}
}

@article{pan2025trinity,
  title={Trinity-rft: A general-purpose and unified framework for reinforcement fine-tuning of large language models},
  author={Pan, Xuchen and Chen, Yanxi and Chen, Yushuo and Sun, Yuchang and Chen, Daoyuan and Zhang, Wenhao and Xie, Yuexiang and Huang, Yilun and Zhang, Yilei and Gao, Dawei and others},
  journal={arXiv preprint arXiv:2505.17826},
  year={2025}
}

@article{chaudhari2025rlhf,
  title={Rlhf deciphered: A critical analysis of reinforcement learning from human feedback for llms},
  author={Chaudhari, Shreyas and Aggarwal, Pranjal and Murahari, Vishvak and Rajpurohit, Tanmay and Kalyan, Ashwin and Narasimhan, Karthik and Deshpande, Ameet and Castro da Silva, Bruno},
  journal={ACM Computing Surveys},
  volume={58},
  number={2},
  pages={1--37},
  year={2025},
  publisher={ACM New York, NY}
}
